\documentclass[12pt]{article}
\newcommand{\filename}{2LM-M\_ANN\_grid\_search\_v8\_A2}
\usepackage{authblk}
\usepackage{amsfonts}
\usepackage{amssymb}
\usepackage{amsmath}
\usepackage{amsthm}
\usepackage{framed}
\usepackage{graphicx}

\usepackage{enumitem}
\usepackage{geometry}
\usepackage{url}
\usepackage{booktabs}

\usepackage{multirow}

\usepackage{mdframed}
\usepackage{mathtools}
\usepackage{algorithm}  
\usepackage{algorithmic}

\newcommand{\Dtrain}{D_\mathrm{train}}  
\newcommand{\Dtest}{D_\mathrm{test}}  
 
\newcommand{\itstp}{\mathrm{ite}^\mathrm{stp}} 
\newcommand{\rstp}{\rho^\mathrm{stp}}

\newcommand{\lf}{\mathrm{lf}}

\newcommand{\newone}{\marginpar{\bf NEW!}}

\newcommand{\eledeg}{\mathrm{eledeg}}  
\newcommand{\eledegC}{\mathrm{eledeg}_\mathrm{C}}   
\newcommand{\eledegT}{\mathrm{eledeg}_\mathrm{T}}   
\newcommand{\eledegF}{\mathrm{eledeg}_\mathrm{F}}   
\newcommand{\eledegX}{\mathrm{eledeg}_\mathrm{X}}   

\newcommand{\vion}{\mathrm{v}_\mathrm{ion}}  

\newcommand{\ttH}{{\tt H}}  
\newcommand{\ttC}{{\tt C}}  
\newcommand{\ttO}{{\tt O}}  
\newcommand{\ttN}{{\tt N}}  
\newcommand{\ttP}{{\tt P}}  
\newcommand{\ttF}{{\tt F}}  
\newcommand{\ttCl}{{\tt Cl}}  
\newcommand{\ttS}{{\tt S}}  
\newcommand{\ttSi}{{\tt Si}}  
\newcommand{\ttPb}{{\tt Pb}}  
\newcommand{\ttBr}{{\tt Br}}

\newcommand{\oH}{\overline{{\tt H}}}  

\newcommand{\Z}{\mathbb{Z}}  
  
\newcommand{\C}{\mathbb{C}}  
\newcommand{\Co}{\mathbb{C}}

\newcommand{\anC}{\langle \mathbb{C} \rangle}  
\newcommand{\anpsi}{\langle \psi \rangle}  

\newcommand{\VH}{V_{\tt H}}

\newcommand{\R}{\mathbb{R}} 
\newcommand{\RK}{\mathbb{R}^K} 
\newcommand{\RKw}{\mathbb{R}^{K+1}}

\newcommand{\deghyd}{\deg^\mathrm{hyd}}

\newcommand{\FrC}{\mathcal{F}^\mathrm{C}} 
\newcommand{\FrT}{\mathcal{F}^\mathrm{T}} 
\newcommand{\FrF}{\mathcal{F}^\mathrm{F}} 
\newcommand{\FrX}{\mathcal{F}^\mathrm{X}}  

\newcommand{\dcp}{\mathrm{dcp}}

\newcommand{\Vleaf}{V_\mathrm{leaf}} 
\newcommand{\Eleaf}{E_\mathrm{leaf}} 

\newcommand{\sint}{\sigma_\mathrm{int}} 
\newcommand{\sce}{\sigma_\mathrm{ce}}

\newcommand{\Ez}{E_{(0/1)}}
\newcommand{\Ew}{E_{(\geq 1)}}
\newcommand{\Et}{E_{(\geq 2)}}
\newcommand{\Eew}{E_{(=1)}}

\newcommand{\Iz}{I_{(0/1)}}
\newcommand{\Iw}{I_{(\geq 1)}}
\newcommand{\It}{I_{(\geq 2)}}
\newcommand{\Iew}{I_{(=1)}}

\newcommand{\Gac}{\Gamma_\mathrm{ac}} 
\newcommand{\Gacs}{\Gamma_\mathrm{ac,<}} 
\newcommand{\Gace}{\Gamma_\mathrm{ac,=}} 
\newcommand{\Gacl}{\Gamma_\mathrm{ac,>}}

\newcommand{\tGacC}{\widetilde{\Gamma}_\mathrm{ac}^\mathrm{C}}  
\newcommand{\tGacT}{\widetilde{\Gamma}_\mathrm{ac}^\mathrm{T}} 
\newcommand{\tGacF}{\widetilde{\Gamma}_\mathrm{ac}^\mathrm{F}}  
\newcommand{\tGacCT}{\widetilde{\Gamma}_\mathrm{ac}^\mathrm{CT}}  
\newcommand{\tGacTC}{\widetilde{\Gamma}_\mathrm{ac}^\mathrm{TC}}  
\newcommand{\tGacCF}{\widetilde{\Gamma}_\mathrm{ac}^\mathrm{CF}}  
\newcommand{\tGacTF}{\widetilde{\Gamma}_\mathrm{ac}^\mathrm{TF}}

\newcommand{\tLdgX}{\widetilde{\Lambda}_\mathrm{dg}^\mathrm{X}}   
\newcommand{\tLdgC}{\widetilde{\Lambda}_\mathrm{dg}^\mathrm{C}}   
\newcommand{\tLdgT}{\widetilde{\Lambda}_\mathrm{dg}^\mathrm{T}}   
\newcommand{\tLdgF}{\widetilde{\Lambda}_\mathrm{dg}^\mathrm{F}}

\newcommand{\tGecC}{\widetilde{\Gamma}_\mathrm{ec}^\mathrm{C}}  
\newcommand{\tGecT}{\widetilde{\Gamma}_\mathrm{ec}^\mathrm{T}} 
\newcommand{\tGecF}{\widetilde{\Gamma}_\mathrm{ec}^\mathrm{F}}  
\newcommand{\tGecCT}{\widetilde{\Gamma}_\mathrm{ec}^\mathrm{CT}}  
\newcommand{\tGecTC}{\widetilde{\Gamma}_\mathrm{ec}^\mathrm{TC}}  
\newcommand{\tGecCF}{\widetilde{\Gamma}_\mathrm{ec}^\mathrm{CF}}  
\newcommand{\tGecTF}{\widetilde{\Gamma}_\mathrm{ec}^\mathrm{TF}}

\newcommand{\typ}{\mathrm{t}}

\newcommand{\x}{x}%{\pmb{x}}

\newcommand{\ta}{{\tt a}}
\newcommand{\tb}{{\tt b}}
\newcommand{\Ldg}{\Lambda_{\mathrm{dg}}}

\newcommand{\fc}{\mathrm{fc}} 
\newcommand{\betar}{\beta_\mathrm{r}}

\newcommand{\val}{\mathrm{val}}

\newcommand{\inte}{\mathrm{int}}

\newcommand{\F}{\mathcal{F}}

\newcommand{\T}{\mathcal{T}}

\newcommand{\nint}{\mathrm{n}^\mathrm{int}}

\newcommand{\h}{\mathrm{ht}}

\newcommand{\cs}{\mathrm{cs}}
\newcommand{\ch}{\mathrm{ch}}

\newcommand{\dg}{\mathrm{dg}}

\newcommand{\na}{\mathrm{na}}
 
\newcommand{\naX}{\mathrm{na}_\mathrm{X}}

\newcommand{\naC}{\mathrm{na}_\mathrm{C}}
\newcommand{\naT}{\mathrm{na}_\mathrm{T}}
\newcommand{\naF}{\mathrm{na}_\mathrm{F}}

\newcommand{\ecX}{\mathrm{ec}_\mathrm{X}}

\newcommand{\ecC}{\mathrm{ec}_\mathrm{C}}
\newcommand{\ecT}{\mathrm{ec}_\mathrm{T}}
\newcommand{\ecF}{\mathrm{ec}_\mathrm{F}}
\newcommand{\ecCT}{\mathrm{ec}_\mathrm{CT}}
\newcommand{\ecTC}{\mathrm{ec}_\mathrm{TC}}
\newcommand{\ecTF}{\mathrm{ec}_\mathrm{TF}}
\newcommand{\ecCF}{\mathrm{ec}_\mathrm{CF}}
 
\newcommand{\acX}{\mathrm{ac}_\mathrm{X}}

\newcommand{\acC}{\mathrm{ac}_\mathrm{C}}
\newcommand{\acT}{\mathrm{ac}_\mathrm{T}}
\newcommand{\acF}{\mathrm{ac}_\mathrm{F}}
\newcommand{\acCT}{\mathrm{ac}_\mathrm{CT}}
\newcommand{\acTC}{\mathrm{ac}_\mathrm{TC}}
\newcommand{\acTF}{\mathrm{ac}_\mathrm{TF}}
\newcommand{\acCF}{\mathrm{ac}_\mathrm{CF}}

\newcommand{\bdX}{\mathrm{bd}_\mathrm{X}}

\newcommand{\bdC}{\mathrm{bd}_\mathrm{C}}
\newcommand{\bdT}{\mathrm{bd}_\mathrm{T}}
\newcommand{\bdF}{\mathrm{bd}_\mathrm{F}}
\newcommand{\bdCT}{\mathrm{bd}_\mathrm{CT}}
\newcommand{\bdTC}{\mathrm{bd}_\mathrm{TC}}
\newcommand{\bdTF}{\mathrm{bd}_\mathrm{TF}}
\newcommand{\bdCF}{\mathrm{bd}_\mathrm{CF}}

\newcommand{\ns}{\mathrm{ns}}

\newcommand{\ec}{\mathrm{ec}}
\newcommand{\ac}{\mathrm{ac}}

\newcommand{\bl}{\mathrm{bl}}

\newcommand{\bd}{\mathrm{bd}}

\newcommand{\UB}{\mathrm{UB}}
\newcommand{\LB}{\mathrm{LB}}

\newcommand{\ex}{\mathrm{ex}}

\newcommand{\GC}{G_\mathrm{C}}

\newcommand{\mC}{m_\mathrm{C}}

\newcommand{\hC}{h^\mathrm{C}}
\newcommand{\hT}{h^\mathrm{T}} 
 
\newcommand{\hX}{h^\mathrm{X}}

\newcommand{\VF}{V_\mathrm{F}}
\newcommand{\VT}{V_\mathrm{T}}
\newcommand{\VC}{V_\mathrm{C}} 
 
\newcommand{\VX}{V_\mathrm{X}}

\newcommand{\ET}{E_\mathrm{T}}
\newcommand{\EC}{E_\mathrm{C}}

\newcommand{\EF}{E_\mathrm{F}}
\newcommand{\ECT}{E_\mathrm{CT}}
\newcommand{\ETC}{E_\mathrm{TC}}
\newcommand{\ETF}{E_\mathrm{TF}}

\newcommand{\ECF}{E_\mathrm{CF}}

\newcommand{\EX}{E_\mathrm{X}}

\newcommand{\vT}{{v^\mathrm{T}}}
\newcommand{\vC}{{v^\mathrm{C}}} 
  
\newcommand{\vX}{{v^\mathrm{X}}}

\newcommand{\vF}{{v^\mathrm{F}}}

\newcommand{\eF}{{e^\mathrm{F}}}
\newcommand{\eT}{{e^\mathrm{T}}}
\newcommand{\eC}{{e^\mathrm{C}}} 
 
\newcommand{\eX}{{e^\mathrm{X}}}

\newcommand{\eCF}{{e^\mathrm{CF}}}

\newcommand{\eCT}{{e^\mathrm{CT}}}
\newcommand{\eTC}{{e^\mathrm{TC}}}
\newcommand{\eTF}{{e^\mathrm{TF}}}

\newcommand{\tT}{{t_\mathrm{T}}}
\newcommand{\tC}{{t_\mathrm{C}}} 
\newcommand{\tF}{{t_\mathrm{F}}} 
 
\newcommand{\tX}{{t_\mathrm{X}}}

\newcommand{\IC}{{I_\mathrm{C}}}

\newcommand{\degCint}{{\deg_\mathrm{C}^\mathrm{int}}}
\newcommand{\degTint}{{\deg_\mathrm{T}^\mathrm{int}}}
\newcommand{\degFint}{{\deg_\mathrm{F}^\mathrm{int}}}
\newcommand{\degXint}{{\deg_\mathrm{X}^\mathrm{int}}}

\newcommand{\hyddeg}{\mathrm{hyddeg}}

\newcommand{\hyddegX}{\mathrm{hyddeg}^\mathrm{X}}

\newcommand{\degCex}{{\deg_\mathrm{C}^\mathrm{ex}}}
\newcommand{\degTex}{{\deg_\mathrm{T}^\mathrm{ex}}}
\newcommand{\degFex}{{\deg_\mathrm{F}^\mathrm{ex}}}
\newcommand{\degXex}{{\deg_\mathrm{X}^\mathrm{ex}}}
\newcommand{\degF}{{\deg^\mathrm{F}}}
\newcommand{\degT}{{\deg^\mathrm{T}}}
\newcommand{\degC}{{\deg^\mathrm{C}}} 
 
\newcommand{\degX}{{\deg^\mathrm{X}}}

\newcommand{\degCT}{\deg_\mathrm{CT}}
\newcommand{\degTC}{\deg_\mathrm{TC}}

\newcommand{\degCTT}{\deg^\mathrm{CT}_\mathrm{T}}
\newcommand{\degTCT}{\deg^\mathrm{TC}_\mathrm{T}}
\newcommand{\degCFF}{\deg^\mathrm{CF}_\mathrm{F}}
\newcommand{\degTFF}{\deg^\mathrm{TF}_\mathrm{F}}

\newcommand{\tldgC}{{\widetilde{\deg}_\mathrm{C}} }

\newcommand{\cF}{{c_\mathrm{F}}}

\newcommand{\kC}{{k_\mathrm{C}}}

\newcommand{\chiF}{{\chi^\mathrm{F}}} 
\newcommand{\dclrF}{\delta_{\chi}^\mathrm{F}}
\newcommand{\clrF}{\mathrm{clr}^{\mathrm{F}}}

\newcommand{\chiT}{{\chi^\mathrm{T}}}
\newcommand{\dclrT}{\delta_{\chi}^\mathrm{T}}
\newcommand{\clrT}{\mathrm{clr}^{\mathrm{T}}}

\newcommand{\tail}{\mathrm{tail}} 
\newcommand{\hd}{\mathrm{head}}

\newcommand{\dlfrF}{\delta_\mathrm{fr}^\mathrm{F}}

\newcommand{\dlfrC}{\delta_\mathrm{fr}^\mathrm{C}} 
 
\newcommand{\dlfrX}{\delta_\mathrm{fr}^\mathrm{X}}
 
\newcommand{\ddgF}{\delta_\mathrm{dg}^\mathrm{F}}
\newcommand{\ddgT}{\delta_\mathrm{dg}^\mathrm{T}}
\newcommand{\ddgC}{\delta_\mathrm{dg}^\mathrm{C}} 
 
\newcommand{\ddgX}{\delta_\mathrm{dg}^\mathrm{X}}

\newcommand{\ddgFint}{\delta_\mathrm{dg,F}^\mathrm{int}}
\newcommand{\ddgTint}{\delta_\mathrm{dg,T}^\mathrm{int}}
\newcommand{\ddgCint}{\delta_\mathrm{dg,C}^\mathrm{int}}  
\newcommand{\ddgXint}{\delta_\mathrm{dg,X}^\mathrm{int}}

\newcommand{\bF}{\beta^\mathrm{F}}
\newcommand{\bT}{\beta^\mathrm{T}}
\newcommand{\bC}{\beta^\mathrm{C}} 
\newcommand{\bX}{\beta^\mathrm{X}}
  
\newcommand{\bCT}{\beta^\mathrm{CT}}
\newcommand{\bTC}{\beta^\mathrm{TC}} 
\newcommand{\bTF}{\beta^\mathrm{TF}} 
\newcommand{\bCF}{\beta^\mathrm{CF}} 
\newcommand{\bXF}{\beta^\mathrm{XF}} 
\newcommand{\bsF}{\beta^{*\mathrm{F}}}

\newcommand{\bFex}{\beta^\mathrm{F}_\mathrm{ex}} 
\newcommand{\bTex}{\beta^\mathrm{T}_\mathrm{ex}} 
\newcommand{\bCex}{\beta^\mathrm{C}_\mathrm{ex}} 
\newcommand{\bXex}{\beta^\mathrm{X}_\mathrm{ex}}

\newcommand{\delbF}{\delta_{\beta}^\mathrm{F}}
\newcommand{\delbT}{\delta_{\beta}^\mathrm{T}}
\newcommand{\delbC}{\delta_{\beta}^\mathrm{C}}

\newcommand{\delbCT}{\delta_{\beta}^\mathrm{CT}}
\newcommand{\delbTC}{\delta_{\beta}^\mathrm{TC}}

\newcommand{\delbsF}{\delta_{\beta}^{*\mathrm{F}}} 
\newcommand{\delbX}{\delta_{\beta}^\mathrm{X}}

\newcommand{\aF}{{\alpha}^\mathrm{F}}
\newcommand{\aT}{{\alpha}^\mathrm{T}}
\newcommand{\aC}{{\alpha}^\mathrm{C}}  
 
\newcommand{\aX}{{\alpha}^\mathrm{X}}
\newcommand{\aCT}{{\alpha}^\mathrm{CT}}
\newcommand{\aTC}{{\alpha}^\mathrm{TC}}

\newcommand{\aCF}{{\alpha}^\mathrm{CF}}  
\newcommand{\aTF}{{\alpha}^\mathrm{TF}}

\newcommand{\delaC}{\delta_\mathrm{\alpha}^{\mathrm{C}}}
\newcommand{\delaT}{\delta_\mathrm{\alpha}^{\mathrm{T}}}
\newcommand{\delaF}{\delta_\mathrm{\alpha}^{\mathrm{F}}}
\newcommand{\delaX}{\delta_\mathrm{\alpha}^{\mathrm{X}}}

\newcommand{\dlnsF}{\delta_{\mathrm{ns}}^\mathrm{F}}
\newcommand{\dlnsT}{\delta_{\mathrm{ns}}^\mathrm{T}}
\newcommand{\dlnsC}{\delta_{\mathrm{ns}}^\mathrm{C}} 
\newcommand{\dlnsX}{\delta_{\mathrm{ns}}^\mathrm{X}}

\newcommand{\dlacF}{\delta_{\mathrm{ac}}^\mathrm{F}}
\newcommand{\dlacT}{\delta_{\mathrm{ac}}^\mathrm{T}}
\newcommand{\dlacC}{\delta_{\mathrm{ac}}^\mathrm{C}}

\newcommand{\dlacCT}{\delta_{\mathrm{ac}}^\mathrm{CT}}
\newcommand{\dlacTC}{\delta_{\mathrm{ac}}^\mathrm{TC}}

\newcommand{\dlacCF}{\delta_{\mathrm{ac}}^\mathrm{CF}} 
\newcommand{\dlacTF}{\delta_{\mathrm{ac}}^\mathrm{TF}} 
\newcommand{\dlacX}{\delta_{\mathrm{ac}}^\mathrm{X}}

\newcommand{\DlacFp}{\Delta_{\mathrm{ac}}^\mathrm{F+}}
\newcommand{\DlacTp}{\Delta_{\mathrm{ac}}^\mathrm{T+}}
\newcommand{\DlacCp}{\Delta_{\mathrm{ac}}^\mathrm{C+}}

\newcommand{\DlacCTp}{\Delta_{\mathrm{ac}}^\mathrm{CT+}}
\newcommand{\DlacTCp}{\Delta_{\mathrm{ac}}^\mathrm{TC+}}

\newcommand{\DlacCFp}{\Delta_{\mathrm{ac}}^\mathrm{CF+}} 
\newcommand{\DlacTFp}{\Delta_{\mathrm{ac}}^\mathrm{TF+}} 
\newcommand{\DlacXp}{\Delta_{\mathrm{ac}}^\mathrm{X+}}

\newcommand{\DlacFm}{\Delta_{\mathrm{ac}}^\mathrm{F-}}
\newcommand{\DlacTm}{\Delta_{\mathrm{ac}}^\mathrm{T-}}
\newcommand{\DlacCm}{\Delta_{\mathrm{ac}}^\mathrm{C-}}

\newcommand{\DlacCTm}{\Delta_{\mathrm{ac}}^\mathrm{CT-}}
\newcommand{\DlacTCm}{\Delta_{\mathrm{ac}}^\mathrm{TC-}}

\newcommand{\DlacCFm}{\Delta_{\mathrm{ac}}^\mathrm{CF-}} 
\newcommand{\DlacTFm}{\Delta_{\mathrm{ac}}^\mathrm{TF-}} 
\newcommand{\DlacXm}{\Delta_{\mathrm{ac}}^\mathrm{X-}}

\newcommand{\dlecF}{\delta_{\mathrm{ec}}^\mathrm{F}}
\newcommand{\dlecT}{\delta_{\mathrm{ec}}^\mathrm{T}}
\newcommand{\dlecC}{\delta_{\mathrm{ec}}^\mathrm{C}}

\newcommand{\dlecCTC}{\delta_{\mathrm{ec,C}}^\mathrm{CT}}

\newcommand{\dlecTCC}{\delta_{\mathrm{ec,C}}^\mathrm{TC}}

\newcommand{\dlecCFC}{\delta_{\mathrm{ec,C}}^\mathrm{CF}}

\newcommand{\dlecTFT}{\delta_{\mathrm{ec,T}}^\mathrm{TF}} 
 
\newcommand{\dlecX}{\delta_{\mathrm{ec}}^\mathrm{X}}

\newcommand{\DlecFp}{\Delta_{\mathrm{ec}}^\mathrm{F+}}
\newcommand{\DlecTp}{\Delta_{\mathrm{ec}}^\mathrm{T+}}
\newcommand{\DlecCp}{\Delta_{\mathrm{ec}}^\mathrm{C+}}

\newcommand{\DlecCTp}{\Delta_{\mathrm{ec}}^\mathrm{CT+}}
\newcommand{\DlecTCp}{\Delta_{\mathrm{ec}}^\mathrm{TC+}}

\newcommand{\DlecCFp}{\Delta_{\mathrm{ec}}^\mathrm{CF+}} 
\newcommand{\DlecTFp}{\Delta_{\mathrm{ec}}^\mathrm{TF+}} 
\newcommand{\DlecXp}{\Delta_{\mathrm{ec}}^\mathrm{X+}}

\newcommand{\DlecFm}{\Delta_{\mathrm{ec}}^\mathrm{F-}}
\newcommand{\DlecTm}{\Delta_{\mathrm{ec}}^\mathrm{T-}}
\newcommand{\DlecCm}{\Delta_{\mathrm{ec}}^\mathrm{C-}}

\newcommand{\DlecCTm}{\Delta_{\mathrm{ec}}^\mathrm{CT-}}
\newcommand{\DlecTCm}{\Delta_{\mathrm{ec}}^\mathrm{TC-}}

\newcommand{\DlecCFm}{\Delta_{\mathrm{ec}}^\mathrm{CF-}} 
\newcommand{\DlecTFm}{\Delta_{\mathrm{ec}}^\mathrm{TF-}} 
\newcommand{\DlecXm}{\Delta_{\mathrm{ec}}^\mathrm{X-}}

\begin{document} 

\begin{center}
   {\Large\bf 
    Molecular Design Based on 
   Artificial Neural Networks, Integer Programming and Grid Neighbor Search} 
\end{center}

\begin{center} 
Naveed Ahmed Azam$^1$, %\orcidID{0000-0002-7941-3419}
Jianshen Zhu$^1$, 
Kazuya Haraguchi$^{1}$, %\orcidID{0000-0002-2479-3135}
Liang Zhao$^2$, %\orcidID{0000-0003-0869-7896}
Hiroshi Nagamochi$^1$ %\orcidID{0000-0002-8332-1517} 
 and  
 Tatsuya Akutsu$^3$ %\orcidID{0000-0001-9763-797X}
\end{center} 
{\small 
 $^1$  Department of Applied Mathematics and Physics, Kyoto University, 
 Kyoto 606-8501, Japan\\
%  \email{\{ zhujs, azam, haraguchi, nag\}@amp.i.kyoto-u.ac.jp}, \\
$^2$   Graduate School of Advanced Integrated Studies in Human Survavibility
     (Shishu-Kan),   Kyoto University, Kyoto 606-8306, Japan \\
%\email{liang@gsais.kyoto-u.ac.jp}  \and
$^3$   Bioinformatics Center,  Institute for Chemical Research, 
  Kyoto University, Uji 611-0011, Japan 
% \email{takutsu@kuicr.kyoto-u.ac.jp} 
}

\begin{quote}  
{\bf Abstract}\\  
% (1)~Background.
A novel framework has recently been proposed for designing 
the molecular structure of chemical compounds
with a desired chemical property  
using both artificial neural networks 
 and mixed integer linear programming.  
In the framework, a chemical graph with a target chemical value is inferred
as a feasible solution of a mixed integer linear program
that represents a prediction function and other requirements
on the structure of graphs.
% (2)~Method.
In this paper, we propose a procedure for generating
other feasible solutions of the mixed integer linear program
by searching the neighbor of output chemical graph
in a search space. 
The procedure is combined in the framework as a new building block. 
% (3)~Results and Conclusion.
The results of our computational experiments suggest that  
 the proposed method can generate an additional number of new chemical graphs  
with  up to 50 non-hydrogen atoms.

\noindent 
{\bf Keywords: } Machine Learning,  Integer Programming,
Cheminformatics, Materials Informatics,
QSAR/QSPR, Molecular Design. 

% QSAR/QSPR,  Molecular Design, 
%    Artificial Neural Network, Mixed  Integer Linear Programming, 
%   Enumeration of Graphs  

% \noindent {\bf  Mathematics Subject Classification: } 
% Primary  
% 05C92,  % Chemical graph theory 
% 92E10, % Molecular structure (graph-theoretic methods, methods of differential topology, etc.)
% Secondary
% 05C30, % Enumeration in graph theory
% 68T07, % Artificial neural networks and deep learning
% 90C11,  % Mixed integer programming
% 92-04 %Software, source code, etc. for problems pertaining to biology
% 05C85 Graph algorithms 
\end{quote}

\section{Introduction}\label{sec:introduction}

\noindent {\bf Background~}
Analysis of chemical compounds is one of the important applications of
intelligent computing.
Indeed, various machine learning methods have been applied to
the prediction of chemical activities from their structural data,
where such a problem is often referred to as
\emph{quantitative structure activity relationship} (QSAR)
\cite{Lo18,Tetko20}.
Recently, neural networks and deep-learning technologies have extensively
been applied to QSAR \cite{Ghasemi18}.

In addition to QSAR, extensive studies have been done on
inverse quantitative structure activity relationship
(inverse QSAR), which seeks for chemical structures having
desired chemical activities under some constraints.
Since it is difficult to directly handle chemical structures
in both QSAR and inverse QSAR,
chemical compounds are usually represented 
as vectors of real or integer numbers,
which are often called \emph{descriptors} in chemoinformatics and
correspond to \emph{feature vectors} in machine learning.
One major approach in inverse QSAR is 
to infer feature vectors from given chemical activities and constraints
and then reconstruct chemical structures from these feature
vectors~\cite{Miyao16,Ikebata17,Rupakheti15},
where chemical structures are usually treated as undirected graphs.
However, the  reconstruction itself is a challenging task
because the number of possible chemical graphs is huge.
For example, 
chemical graphs with up to 30 atoms (vertices)
{\tt C}, {\tt N}, {\tt O}, and  {\tt S}
may  exceed~$10^{60}$~\cite{BMG96}. 
Indeed, it is NP-hard to infer a chemical graph from a given feature vector
except for some simple cases~\cite{AFJS12}.  
Due to this inherent difficulty, most existing methods for  inverse QSAR
do not guarantee optimal or exact solutions.

As a new approach,
extensive studies have recently been done for inverse QSAR using 
\emph{artificial neural networks} (ANNs),
especially using graph convolutional networks~\cite{Kipf16}.
For example, recurrent neural networks~\cite{Segler18,Yang17}, 
variational autoencoders~\cite{Gomez18}, 
grammar variational autoencoders~\cite{Kusner17},
generative adversarial networks~\cite{DeCao18},
and invertible flow models~\cite{Madhawa19,Shi20}
have been applied.
However, these methods do not yet guarantee optimal or exact solutions.  

\begin{figure}[!ht]  \begin{center}
\includegraphics[width=.77\columnwidth]{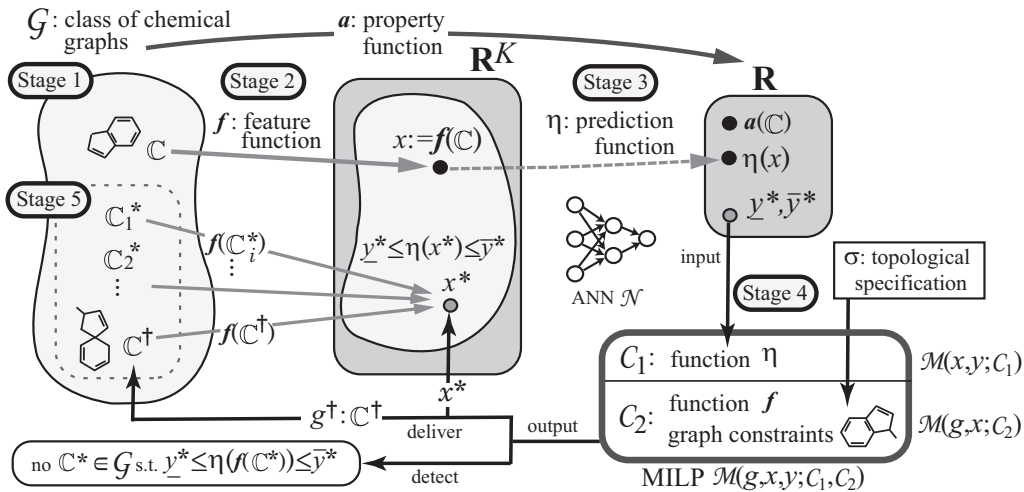}
\end{center} \caption{An illustration of a framework for inferring
a set of chemical graphs $\Co^*$.   } 
\label{fig:framework}  \end{figure}    

\smallskip
\noindent {\bf Framework~}
Akutsu and Nagamochi~\cite{AN19} proved that  
the computation process of a given ANN can be simulated
with a mixed integer linear programming (MILP).
Based on this,
a novel  framework for inferring chemical graphs has been developed
and revised~\cite{ACZSNA20,ZAHZNA21}, 
as illustrated in Figure~\ref{fig:framework}. 
It constructs a prediction function in the first phase and
infers a chemical graph in the second phase. 
The first phase of the framework consists of three stages.
In Stage~1, we choose a chemical property $\pi$ and a class $\mathcal{G}$ 
of graphs, where a property function
$a$ is defined so that $a(\Co)$ is the value of $\pi$  for a compound $\Co\in \mathcal{G}$,
and collect a data set $D_{\pi}$ of chemical graphs in  $\mathcal{G}$ 
such that $a(\Co)$ is available for every $\Co \in D_{\pi}$.
In Stage~2, we introduce a feature function $f: \mathcal{G}\to \mathbb{R}^K$ 
for a positive integer $K$.    
In Stage~3, we construct a prediction function $\eta$ 
with an ANN $\mathcal{N}$ that,  
given a   vector  $x\in \mathbb{R}^K$, 
returns a value $y=\eta(x)\in \mathbb{R}$    
so that $\eta(f(\Co))$ serves as a predicted value
to the real value $a(\Co)$ of $\pi$ for each $\Co\in D_\pi$.  
Given two reals $\underline{y}^*$ and $\overline{y}^*$
 as an interval  for a  target chemical value,
the  second phase infers  chemical graphs $\Co^*$
with $\underline{y}^*\leq \eta(f(\C^*))\leq \overline{y}^*$ in the next two stages. 
We have obtained a  feature function $f$ and a  prediction function $\eta$
and call an additional constraint on the substructures of target chemical graphs 
a {\em topological specification}. 
In Stage~4, we prepare the following two  MILP formulations: 
\begin{enumerate}[nosep,  leftmargin=*]
\item[-]
 MILP $\mathcal{M}(x,y;\mathcal{C}_1)$
with a set $\mathcal{C}_1$ of linear constraints on variables $x$ and $y$
(and some other auxiliary variables) 
  simulates the  process of computing $y:=\eta(x)$ from a vector $x$; and
\item[-]
 MILP $\mathcal{M}(g,x;\mathcal{C}_2)$
with a set $\mathcal{C}_2$ of linear constraints on  variable  $x$ and
 a variable vector  $g$ that represents a chemical graph $\Co$
(and some other auxiliary variables)  
  simulates the  process of computing $x:=f(\Co)$ from a chemical graph $\Co$
and chooses a chemical graph $\Co$ that satisfies the given topological specification
$\sigma$. 
\end{enumerate} 
Given an interval with boundaries $\underline{y}^*,\overline{y}^* \in \mathbb{R}$,
 we solve the combined
MILP $\mathcal{M}(g,x,y;\mathcal{C}_1,\mathcal{C}_2)$
to find  a feature vector $x^*\in \mathbb{R}^K$
 and a chemical graph $\Co^{\dagger}$  with the specification
$\sigma$ such that $f(\Co^\dagger)=x^*$ and  
$\underline{y}^*\leq \eta(x^*) \leq \overline{y}^*$
(where if the MILP instance is infeasible then this suggests that there 
does  not exist
such a desired chemical graph).
In Stage~5,  we generate other  chemical graphs $\Co^*$
such that $\underline{y}^*\leq \eta(f(\C^*))\leq \overline{y}^*$
 based on the output chemical graph $\Co^\dagger$.

MILP formulations required in Stage~4 have been designed   
for chemical compounds with 
%cycle index 0 (i.e., acyclic) \cite{ZZCSNA20,AZSSSZNA20},
%cycle index 1~\cite{IAWSNA20} and 
cycle index at most 2~\cite{ZCSNA20}.
Afterwards, a modeling of chemical compounds together with 
an MILP formulation has been improved so that
a chemical compound with any graph structure can treated
(see  Shi et~al.~\cite{SZAHZNA21}). 
Not only ANNs but also other machine learning methods
have been used to construct a prediction function $\eta$ in Stage~3
recently. 
Tanaka et~al.~\cite{TZAHZNA21} 
 (resp.,  Zhu~et~al.~\cite{ZAHZNA21}) used a decision tree
  (resp.,  linear regression)
to construct a prediction function $\eta$ in Stage~3 in the framework
and derived an MILP $\mathcal{M}(x,y;\mathcal{C}_1)$ 
that simulates the computation process of  a decision tree
 (resp.,  linear regression).

\smallskip
\noindent {\bf Contribution~}
In the current framework,  a chemical graph $\C^\dagger$ with a desired 
chemical property $\pi$ in Stage~4 is constructed 
as a feasible solution of an MILP $\mathcal{M}(g,x,y;\mathcal{C}_1,\mathcal{C}_2)$
before we generate isomers $\C^*$ of $\C^\dagger$ by an algorithm such as 
the dynamic programming algorithm due to Azam~et~al.~\cite{AZSSSZNA20}. 
In this paper, we design  a procedure of generating solutions
of the same  MILP as a new building block in Stage~4 of the framework of inferring chemical graphs. 
When a feasible solution  $\C^\dagger$ of the MILP is constructed in Stage~4,
we try to find other feasible solutions $\C^*$ of the same MILP by
solving the MILP with additional $p_{\max}$  linear constraints for some integer $p_{\max}$.
For this,  we first prepare arbitrary linear functions $\theta_p: \RK\to \R, p\in [1,p_{\max}]$
and consider a neighbor of $\C^\dagger$ 
 defined by a set of chemical graphs $\C^*$ that satisfy linear constraints 
 $\delta_1\leq |\theta_p(f(\C^*))-\theta_p(f(\C^\dagger))|\leq \delta_2, p\in [1,p_{\max}]$ for 
 small reals $\delta_2>\delta_1>0$.  
By changing the  reals $\delta_2$ and $\delta_1$ systematically, we can search for new solutions
of the same MILP other than $\C^\dagger$.
As a candidate for a function $\theta_p$, we can use a linear prediction 
function $\eta_\tau: \RK\to \R$  
for a different chemical property $\tau$ such as a function $\eta_\tau$ that has been obtained
by linear regression. 
With a linear prediction function $\theta_p=\eta_\tau$, we can 
search for other chemical graphs $\C^*$ by specifying a predicted value of $\C^*$ for the property $\tau$.

We implemented the framework with the new building block 
based on  the two-layered model and the feature function
proposed by  Zhu~et~al.~\cite{ZAHZNA21}. 
We used  the same MILP $\mathcal{M}(g,x;\mathcal{C}_2)$ 
formulation proposed by  Zhu~et~al.~\cite{ZAHZNA21} 
 and omit the details in this paper.  
From the results of our computational experiments,
we observe that 
 the proposed method can generate an additional number of chemical graphs $\C^*$ in Stage~4
with  up to 50 non-hydrogen atoms. 
% Among the data sets tested in the experiment, we use a data of polymers
% for the first time in the framework.

The paper is organized as follows.  
Section~\ref{sec:preliminary} introduces some notions on graphs,
 a modeling of chemical compounds and a choice of descriptors.  
Section~\ref{sec:2LM} reviews  the two-layered model.
Section~\ref{sec:grid_search} introduces a new method of
generating solutions of an MILP in Stage~4. 
Section~\ref{sec:experiment} reports the results on  computational 
experiments conducted for  47 chemical properties such as 
biological half life and boiling point for monomers
and  characteristic ratio and refractive index for polymers. 
Section~\ref{sec:conclude} makes some concluding remarks.   
Some technical details are given in Appendices:   
 Appendix~\ref{sec:descriptor} for  all descriptors in our feature function; 
 Appendix~\ref{sec:specification} for a full description of 
a topological specification; and 
Appendix~\ref{sec:test_instances} for the detail of test instances
used in our computational experiment for Stages~4 and 5.
%Appendix~\ref{sec:full_milp}
%  for the details of   our MILP formulation  $\mathcal{M}(g,x;\mathcal{C}_2)$.
  
% that represents a feature function $f$ and a specification $\sigma$.   
%
% The proposed method/system is available at GitHub {\tt  https://github.com/ku-dml/mol-infer}.

\section{Preliminary}\label{sec:preliminary}%%%%%%%%%

This section  introduces some notions and terminologies on graphs,
  modeling of chemical compounds and our choice of descriptors. 
 
Let $\mathbb{R}$, $\mathbb{R}_+$, $\mathbb{Z}$  and $\mathbb{Z}_+$ 
denote the sets of reals,  non-negative reals, 
integers and non-negative integers, respectively.
For two integers $a$ and $b$, let $[a,b]$ denote the set of 
integers $i$ with $a\leq i\leq b$.
For a vector $x\in \R^p$, the $j$-th entry of $x$ is denoted by $x(j)$.

\bigskip\noindent
{\bf  Graph} 
Given a  graph $G$, let $V(G)$ and $E(G)$ denote the sets
of vertices and edges, respectively.     
For a subset $V'\subseteq V(G)$ (resp., $E'\subseteq E(G))$ of
a graph $G$, 
let $G-V'$ (resp., $G-E'$) denote the graph obtained from $G$
by removing the vertices in $V'$ (resp.,  the edges in $E'$),
where we remove all edges incident to a vertex in $V'$
in $G-V'$. 
An edge subset $E'\subseteq E(G)$ in a connected graph $G$ is called
{\em separating} (resp., {\em non-separating})
if $G-E'$  becomes disconnected (resp., $G-E'$ remains connected). 
The {\em rank}  $\mathrm{r}(G)$ of a graph $G$  is defined to be 
the minimum $|F|$ of an edge subset $F\subseteq E(G)$
such that $G-F$ contains no cycle, where 
$\mathrm{r}(G)=|E(G)|-|V(G)|+1$ for a connected graph $G$.   
Observe that   $\mathrm{r}(G-E')=\mathrm{r}(G)-|E'|$ holds
for any non-separating edge subset $E'\subseteq E(G)$. 
%We call a  graph $G$ with rank $k$ a  {\em rank-$k$  graph}.
An edge $e\in E(G)$ in a connected graph $G$
  is called a {\em bridge} if $\{e\}$ is separating.
%, i.e.,
% $G-e$ consists of two connected graphs $G_i$ containing vertex $u_i$, $i=1,2$.
% A bridge  is called a {\em leaf-edge} if one of the end-vertices is of degree 1.
For a connected cyclic graph $G$, an edge $e$ is called a {\em core-edge} if
it is in a cycle of $G$ or is a bridge $e=u_1u_2$ such that
each of the connected graphs $G_i$, $i=1,2$ of $G-e$ contains a cycle. 
A vertex incident to a core-edge is called a {\em core-vertex} of $G$. 
A path with two end-vertices $u$ and $v$ is called a {\em $u,v$-path}. 
 
 We define a {\em rooted} graph to be
 a graph with a  designated vertex, called a {\em root}. 
%In this paper, we designated at most two vertices as roots, 
%and denote by $\mathrm{Rt}(G)$ the set of roots of $G$.
% We call a graph $G$   {\em rooted} (resp., {\em bi-rooted})
% if $|\mathrm{Rt}(G)|=1$ (resp., $|\mathrm{Rt}(G)|=2$),
% where we call $G$ {\em unrooted} if $\mathrm{Rt}(G)=\emptyset$.
 %
%
 For a graph $G$ possibly with a root,  
 a {\em leaf-vertex} is defined to be a non-root vertex 
 with degree 1. 
We call  the edge $uv$ incident to a leaf vertex $v$ a {\em leaf-edge},
 and denote by $\Vleaf(G)$ and $\Eleaf(G)$
  the sets of leaf-vertices and leaf-edges  in $G$, respectively.
 For a graph  or a rooted graph $G$,
 we define graphs $G_i, i\in \mathbb{Z}_+$ obtained from $G$
 by removing the set of leaf-vertices $i$ times so that
\[ G_0:=G; ~~ G_{i+1}:=G_i - \Vleaf(G_i), \]
where we call a vertex $v$ a {\em tree vertex} if $v\in \Vleaf(G_i)$
for some $i\geq 0$. 
Define the {\em height} $\h(v)$ of each tree vertex $v\in \Vleaf(G_i)$
to be $i$; and 
$\h(v)$ of each non-tree vertex $v$ adjacent to a tree vertex 
to be $\h(u)+1$ for the maximum $\h(u)$ of a tree vertex $u$ adjacent to $v$,
where we do not define height of any non-tree vertex not adjacent to any tree vertex. 
We call a vertex $v$ with $\h(v)=k$ a {\em leaf $k$-branch}.
The {\em height} $\h(T)$ of a rooted tree $T$ is defined
to be the maximum of $\h(v)$ of a vertex $v\in V(T)$. 
For an integer $k\geq 0$, we call a  rooted tree $T$
 {\em $k$-lean} if $T$ has at most one leaf $k$-branch.
For an unrooted cyclic graph $G$, we regard that
the set of non-core-edges in $G$ induces
a collection $\mathcal{T}$ of trees each of which is rooted at a core-vertex,
where we call $G$  {\em $k$-lean} if each of the rooted trees in $\mathcal{T}$ 
is $k$-lean. 
 
\subsection{Modeling of Chemical Compounds}\label{sec:chemical_model}

We review a modeling of chemical compounds introduced 
by  Zhu~et~al.~\cite{ZAHZNA21}. 

To represent a chemical compound, 
we introduce a set  of   chemical elements such as 
  {\tt H} (hydrogen),   
 {\tt C} (carbon), {\tt O} (oxygen), {\tt N} (nitrogen)  and so on.
 To distinguish a chemical element $\ta$ with multiple valences such as {\tt S} (sulfur),
 we denote a chemical element $\ta$ with a valence $i$ by $\ta_{(i)}$,
 where we do not use such a suffix $(i)$ 
 for a chemical element $\ta$ with a unique valence. 
Let $\Lambda$ be a set of chemical elements $\ta_{(i)}$.
For example,  $\Lambda=\{\ttH,  \ttC, \ttO, \ttN, \ttP, \ttS_{(2)}, \ttS_{(4)}, \ttS_{(6)}\}$. 
Let $\val: \Lambda\to [1,6]$ be a valence function.
%\newone
For example, $\val(\ttH)=1$, $\val(\ttC)=4$, $\val(\ttO)=2$, $\val(\ttP)=5$,
$\val(\ttS_{(2)})=2$, $\val(\ttS_{(4)})=4$ and $\val(\ttS_{(6)})=6$.
 For each  chemical element $\ta\in \Lambda$, 
let $\mathrm{mass}(\ta)$  denote the mass   of  $\ta$.
% let $\mathrm{elng}(\ta)$  denote the electronegativity  of  $\ta$,  %(電気陰性度)
% let $\mathrm{ionz}(\ta)$  denote the ionization potential of  $\ta$ and   %  (イオン化電位)
% let $\mathrm{elaf}(\ta)$  denote the electron affinity of  $\ta$.  % (電子親和力) 

A chemical compound  is represented by a {\em chemical graph} defined to be
a tuple $\Co=(H,\alpha,\beta)$  of
  a simple, connected undirected graph $H$ and  
    functions   $\alpha:V(H)\to \Lambda$  and  $\beta: E(H)\to [1,3]$.
The set of atoms and the set of bonds in the compound 
are represented by the vertex set $V(H)$ and the edge set $E(H)$, respectively.
The chemical element assigned to a vertex $v\in V(H)$
is represented by $\alpha(v)$ and 
 the bond-multiplicity  between two adjacent vertices  $u,v\in V(H)$
is represented by $\beta(e)$ of the edge $e=uv\in E(H)$.
We say that two tuples $(H_i,\alpha_i,\beta_i), i=1,2$ are
{\em isomorphic} if they admit an isomorphism $\phi$,
i.e.,  a bijection $\phi: V(H_1)\to V(H_2)$
such that
 $uv\in E(H_1), \alpha_1(u)=\ta, \alpha_1(v)=\tb, \beta_1(uv)=m$
 $\leftrightarrow$  
 $\phi(u)\phi(v) \in E(H_2), \alpha_2(\phi(u))=\ta, 
 \alpha_2(\phi(v))=\tb, \beta_2(\phi(u)\phi(v))=m$. 
 When $H_i$ is rooted at a vertex $r_i,  i=1,2$,
these chemical graphs $(H_i,\alpha_i,\beta_i),  i=1,2$ are
{\em rooted-isomorphic} (r-isomorphic) if 
they admit  an isomorphism $\phi$ such that $\phi(r_1)=r_2$. 
% Chemical rooted trees $T_1$ and $T_5$ 
% in Figure~\ref{fig:example_chemical_graph} are r-isomorphic. 

 For a notational convenience, we  use
 a function $\beta_\Co: V(H)\to [0,12]$ 
 for a chemical graph $\Co=(H,\alpha,\beta)$
  such that $\beta_\Co(u)$ means the sum of bond-multiplicities
 of edges incident to a vertex $u$; i.e., 
\[ \beta_\Co(u) \triangleq \sum_{uv\in E(H) }\beta(uv) 
\mbox{ for each vertex $u\in V(H)$.}\]
For each vertex $u\in V(H)$, 
 define the {\em electron-degree} $\eledeg_\Co(u)$  to be 
%\newone
\[  \eledeg_\Co(u) \triangleq  \beta_\Co(u) - \val(\alpha(u)). \]
For each  vertex $u\in V(H)$, let $\deg_\Co(v)$ denote 
the number of vertices adjacent to $u$ in $\Co$. 
  
  For a chemical   graph  $\Co=(H,\alpha,\beta)$, 
  let  $V_{\ta}(\Co)$, $\ta\in \Lambda$
   denote the set of vertices $v\in V(H)$ such that $\alpha(v)=\ta$ in $\Co$
  and define the {\em hydrogen-suppressed chemical graph} $\anC$ 
to be  the graph obtained from $H$ by
  removing all the vertices $v\in \VH(\Co)$.

\subsection{Evaluating prediction function}\label{sec:coefficient_determination} 

We review the definition of coefficient of determination.

Let $D$ be a data set   of chemical graphs $\C$ with
an observed value $a(\C)\in \R$,
where we denote by $a_i=a(\C_i)$ 
for an indexed graph $\C_i$. 

Let $ f$ be a feature function that maps a chemical graph $\C$
to a vector $ f(\C)\in \RK$
where we denote by $\x_i= f(\C_i)$ 
for an indexed graph $\C_i$. 
For  a prediction function $\eta: \RK\to \R$, 
define an error function 
\[ \mathrm{Err}(\eta;D)  \triangleq 
\sum_{\C_i\in D}(a_i - \eta(f(\C_i)))^2=\sum_{\C_i\in D}(a_i - \eta(\x_i))^2, \]
and define the {\em coefficient of determination}
 $\mathrm{R}^2(\eta,D)$ 
  to be 
\[ \displaystyle{ \mathrm{R}^2(\eta,D)\triangleq 
  1- \frac{\mathrm{Err}(\eta;D) } 
  {\sum_{ \C_i\in D  } (a_i-\widetilde{a})^2}  
%=   1- \frac{\sum_{ \C_i\in D  }(a_i-\eta (f (\C_i)) )^2} 
% {\sum_{ \C_i\in D  } (a_i-\widetilde{a})^2} 
  \mbox{   for  }
   \widetilde{a}= \frac{1}{|D |}\sum_{ \C\in D }a(\C).  } \]

\begin{figure}[h!] \begin{center}
\includegraphics[width=.80\columnwidth]{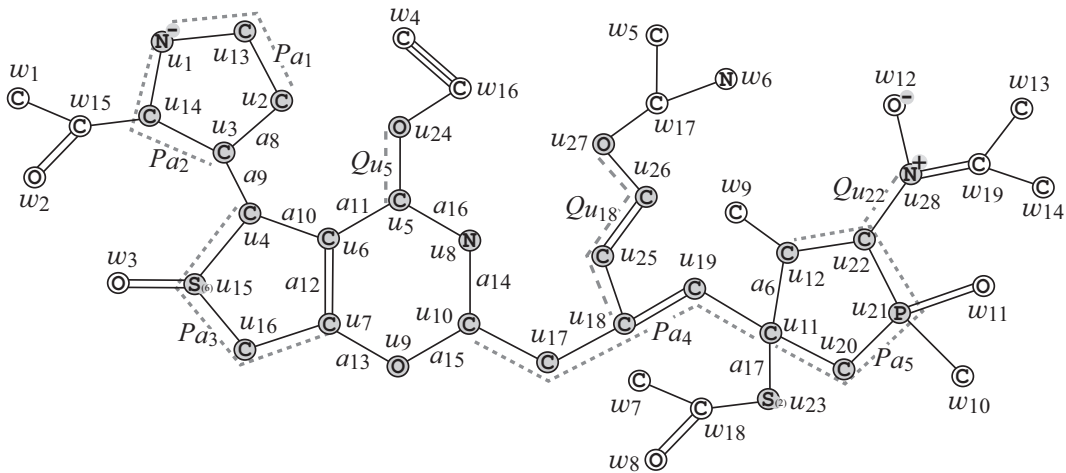}
\end{center} \caption{An illustration of  a hydrogen-suppressed chemical graph  
$\anC$ obtained from a chemical graph $\C$ with $\mathrm{r}(\C)=4$ 
by removing all the %52
 hydrogens, 
where for  ${\rho}=2$,  
$V^\ex(\C)=\{w_i \mid i\in [1,19]\}$ and
$V^\inte(\C)=\{u_i \mid i\in [1,28]\}$.  
 }
\label{fig:example_chemical_graph} \end{figure}

\section{Two-layered Model}\label{sec:2LM}%%%%%%%%%
This section reviews the two-layered model introduced  by 
 Zhu~et~al.~\cite{ZAHZNA21}. 

 Let  $\C=(H,\alpha,\beta)$ be a chemical graph
 and  ${\rho}\geq 1$ be an integer, which we call a {\em branch-parameter}.
 
  A {\em two-layered model} of $\C$ is a partition of
 the hydrogen-suppressed chemical graph $\anC$ into
 an ``interior'' and an ``exterior'' in the following way. 
 We call a vertex $v\in V(\anC)$
   (resp., an edge $e\in E(\anC))$ of   $\C$
   an {\em exterior-vertex} (resp.,    {\em exterior-edge}) if
    $\h(v)< {\rho}$ (resp., $e$ is incident to an  exterior-vertex)
and denote the sets of exterior-vertices and exterior-edges 
by $V^\ex(\C)$ and $E^\ex(\C)$, respectively
and denote  $V^\inte(\C)=V(\anC)\setminus  V^\ex(\C)$ and 
$E^\inte(\C)=E(\anC)\setminus E^\ex(\C)$, respectively.
We call a vertex in $V^\inte(\C)$ (resp.,   an edge in $E^\inte(\C)$) 
   an {\em interior-vertex} (resp.,    {\em interior-edge}). 
 The set  $E^\ex(\C)$ of  exterior-edges forms 
a collection of connected graphs each of which is
regarded as a rooted tree $T$ rooted at 
the vertex $v\in V(T)$ with the maximum $\h(v)$. 
Let $\mathcal{T}^\ex(\anC)$ denote 
the set of these chemical rooted trees in $\anC$. 
The {\em interior} $\C^\inte$ of $\C$ is defined to be the subgraph
 $(V^\inte(\C),E^\inte(\C))$ of $\anC$. 

Figure~\ref{fig:example_chemical_graph}
 illustrates an example of a hydrogen-suppressed chemical graph $\anC$.
For a branch-parameter ${\rho}=2$, 
the interior of  the chemical graph $\anC$ in Figure~\ref{fig:example_chemical_graph} 
is obtained by removing the set of vertices with degree 1 ${\rho}=2$ times; i.e., 
first remove  
the set  $V_1=\{w_1,w_2,\ldots,w_{14}\}$ of vertices of degree 1 in $\anC$ 
and then remove  the set
 $V_2=\{w_{15},w_{16},\ldots,w_{19}\}$ of vertices of degree 1 in $\anC-V_1$,
 where the removed vertices become the exterior-vertices of $\anC$.

%Note that every core-vertex (resp., core-edge) in the graph $H$ is
%an interior-vertex (resp., interior-edge) of $\C$.  

%Figure~\ref{fig:example_chemical_graph} illustrates an example 
%  of a hydrogen-suppressed chemical  graph $\anC$,   such that
% $V^\inte(\C)=\{u_1,u_2,\ldots,u_{28}\}$ and  
% $V^\ex(\C)=\{w_1,w_2,\ldots,w_{19}\}$ 
%  for a branch-parameter ${\rho}=2$. 
  
\begin{figure}[h!] \begin{center}
\includegraphics[width=.84\columnwidth]{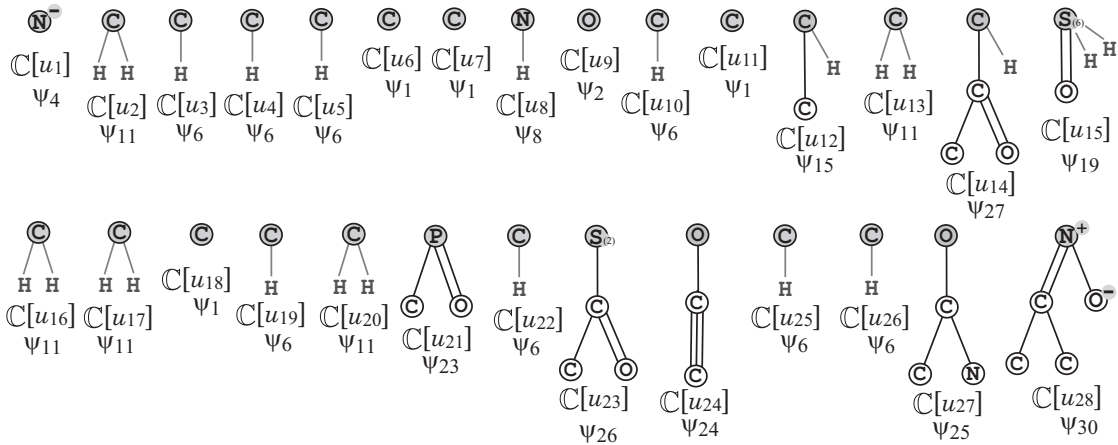}
\end{center} \caption{
The set $\mathcal{T}(\C)$ of  2-fringe-trees  $\C[u_i], i\in [1,28]$ of the example $\C$ with $\anC$ 
in Figure~\ref{fig:example_chemical_graph}, 
where the root of each tree is depicted with a gray circle and
 the hydrogens attached to non-root vertices are omitted in the figure.  
 }
\label{fig:example_fringe-tree} \end{figure}

For each interior-vertex $u\in V^\inte(\C)$,
let $T_u\in \mathcal{T}^\ex(\anC)$ denote the chemical tree rooted at $u$
(where possibly $T_u$ consists of vertex $u$)
and 
define the {\em $\rho$-fringe-tree} $\C[u]$ 
to be  
the chemical rooted tree obtained from $T_u$ by putting back
 the hydrogens originally attached with $T_u$ in $\C$. 
Let $\mathcal{T}(\C)$ denote the set of $\rho$-fringe-trees 
$\C[u], u \in V^\inte(\C)$. 
Figure~\ref{fig:example_fringe-tree}  illustrates
the set  $\mathcal{T}(\C)=\{\C[u_i]\mid i\in [1,28]\}$ of the 2-fringe-trees 
  of the example $\C$ with $\anC$
in Figure~\ref{fig:example_chemical_graph}. 

\smallskip
\noindent {\bf Feature Function~} 
 The feature of an  interior-edge $e=uv\in E^\inte(\C)$ 
 such that $\alpha(u)=\ta$, $\deg_{\anC}(u)=d$, 
 $\alpha(v)=\tb$, $\deg_{\anC}(v)=d'$  and $\beta(e)=m$  is represented by 
 a tuple $(\ta d, \tb d', m)$, which is called the {\em edge-configuration} 
  of the edge $e$, where 
  we call the tuple $(\ta, \tb, m)$ 
 the {\em adjacency-configuration} of the edge $e$. 
 
For an integer $K$, a feature vector $f(\C)$ of a chemical graph $\C$
is defined by a {\em feature function} $f$ that consists of $K$ descriptors. 
We call  $\RK$ {\em  the feature space}.

 Tanaka et~al.~\cite{TZAHZNA21} 
  defined  a feature vector $f(\C)\in \RK$  
to be a combination of the frequency 
of edge-configurations of   the interior-edges  and
the frequency of chemical rooted trees among the set 
of  chemical rooted trees $\C[u]$ over all interior-vertices $u$. 
%In this paper, we introduce the rank and the adjacency-configuration of leaf-edges
%as new descriptors  in a feature vector of a chemical graph. \newone
% a new descriptor for featuring leaf-edges using adjacency-configuration.
 
\smallskip
\noindent {\bf Topological Specification~}   
A topological specification is described
as a set of the following rules proposed by 
Shi et al.~\cite{SZAHZNA21}
and modified by Tanaka et~al.~\cite{TZAHZNA21}:
\begin{enumerate}[nosep]
\item[(i)]
a {\em seed graph} $\GC$ as an  abstract form of  a target chemical graph $\C$;
\item[(ii)]
 a set $\mathcal{F}$ of chemical rooted trees  as candidates
 for a tree  $\C[u]$ rooted at each interior-vertex $u$ in $\C$; 
and 
\item[(iii)]
lower and upper bounds on the number of components 
 in a target chemical graph such as  chemical elements, 
double/triple bonds and the interior-vertices in $\C$. 
\end{enumerate} 

\begin{figure}[h!] \begin{center}
\includegraphics[width=.98\columnwidth]{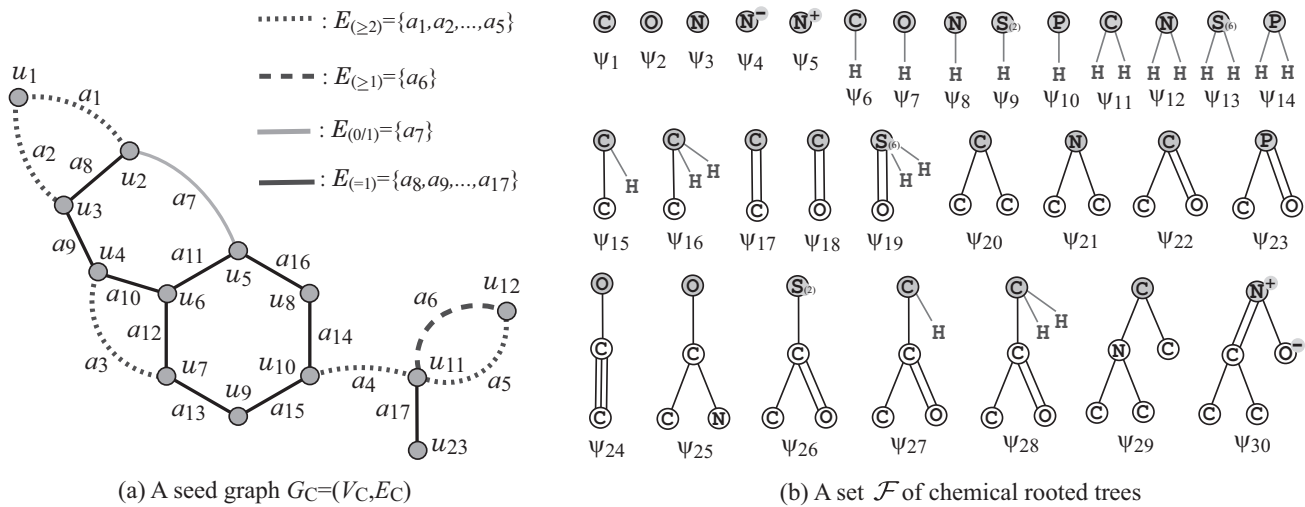}
\end{center} \caption{
(a) An illustration of a seed graph $\GC$ with $\mathrm{r}(\GC)=5$, 
where the vertices in $\VC$ are depicted with gray circles,
the edges in $\Et$ are depicted with dotted lines,
the edges in $\Ew$ are depicted with dashed lines,
the edges in $\Ez$ are depicted with gray bold lines and  
the edges in $\Eew$ are depicted with black solid lines;
(b) A set $\mathcal{F}=\{\psi_1,\psi_2,\ldots,\psi_{30}\}\subseteq
\mathcal{F}(D_\pi)$ of 30 chemical rooted trees
$\psi_i, i\in [1,30]$, where the root of each tree is depicted with a gray circle, 
where  the hydrogens attached to non-root vertices are omitted in the figure.    }
\label{fig:specification_example_1} \end{figure}  

Figure~\ref{fig:specification_example_1}(a) and (b)
 illustrate  examples of  a  seed graph  $\GC$ and 
 a set $\mathcal{F}$ of chemical rooted trees, respectively. 
 Given a seed graph $\GC$, 
 the interior of   a target chemical graph $\C$ is constructed
 from $\GC$ by replacing some edges $a=uv$ 
 with paths $P_a$ between the end-vertices
 $u$ and $v$ and by attaching new paths $Q_v$ to some vertices $v$.  
%
% where some edges in $\GC$ are required to be used in $G$
% and some edge  $a=uv$ in $\GC$ is allowed to be replaced with a path $P_a$
% joining the end-vertices $u$ and $v$.   
For example, a chemical graph $\C$ with $\anC$ 
in Figure~\ref{fig:example_chemical_graph} is constructed
from the seed  graph  $\GC$ in Figure~\ref{fig:specification_example_1}(a)
as follows.
\begin{enumerate}[nosep,  leftmargin=*]
\item[-]
First replace  five edges
 $a_1=u_1 u_{2},  a_2=u_1 u_{3},  a_3=u_4 u_{7}, a_4=u_{10}u_{11}$
and $a_5=u_{11}u_{12}$ in  $\GC$ 
 with new paths  
$P_{a_1}=(u_1,u_{13},u_{2})$, 
$P_{a_2}=(u_{1},u_{14},u_{3})$,
$P_{a_3}=(u_{4},u_{15},u_{16},u_{7})$, 
 $P_{a_4}=(u_{10},u_{17},u_{18},u_{19},u_{11})$ and
 $P_{a_5}=(u_{11},u_{20},u_{21},u_{22},u_{12})$, respectively
 to obtain a subgraph $G_1$ of $\anC$. 
\item[-]
Next attach to this graph  $G_1$ three new paths 
$Q_{u_5}=(u_5,u_{24})$, 
$Q_{u_{18}}=(u_{18},u_{25},u_{26},u_{27})$ and 
$Q_{u_{22}}=(u_{22},u_{28})$
to obtain  
the interior of  $\anC$ in Figure~\ref{fig:example_chemical_graph}.
\item[-]
Finally  attach to the interior   28 trees selected from the set $\mathcal{F}$ 
and assign chemical elements and bond-multiplicities in the interior
to  obtain a chemical graph $\C$ with $\anC$ in Figure~\ref{fig:example_chemical_graph}. 
In Figure~\ref{fig:example_fringe-tree},  
  $\psi_1\in \mathcal{F}$ is selected for $\Co[u_i]$, $i\in\{6,7,11\}$.
    Similarly 
  $\psi_2$  for  $\Co[u_9]$,
  $\psi_4$   for $\Co[u_1]$, 
  $\psi_6$   for $\Co[u_i]$,   
  $i\in\{3,4,5,10,19,22,25,26\}$,
  $\psi_8$   for $\Co[u_8]$, 
  $\psi_{11}$    for $\Co[u_i]$, $i\in\{2,13,16,17,20\}$,
  $\psi_{15}$   for $\Co[u_{12}]$,
   $\psi_{19}$    for $\Co[u_{15}]$,
   $\psi_{23}$    for $\Co[u_{21}]$,
   $\psi_{24}$    for $\Co[u_{24}]$,
   $\psi_{25}$    for $\Co[u_{27}]$, 
   $\psi_{26}$   for $\Co[u_{23}]$, 
   $\psi_{27}$  for $\Co[u_{14}]$     
   and 
    $\psi_{30}$  for $\Co[u_{28}]$. 
\end{enumerate} 

%In (iii), the frequency of chemical elements and the graph size 
%are controlled with  lower and upper bounds on the components 
% in a target chemical graph $\C$. 
%
% See Section~\ref{sec:specification} for more details on the specification. 
%

Our definition of a topological specification is analogous with the one  by 
 Tanaka et~al.~\cite{TZAHZNA21} 
   except for a necessary modification due to the introduction 
   of multiple valences of chemical elements, cations and anions 
(see Appendix~\ref{sec:specification} for a full description of topological specification).

%\clearpage  

\section{Grid Neighbor Search}\label{sec:grid_search}%%%%%%%%%

% For an integer $K\geq 1$, define   a feature space $\RK$. 
%Let $\mathcal{X}=\{x_1,x_2,\ldots,x_m\}$ be a set of feature vectors $x\in \RK$
%and let $a_i\in \R$ be a real assigned to a feature vector $x_i$.
This section introduces a procedure of generating solutions
of an MILP as a new building block of the framework of inferring chemical graphs.

For a notational convenience, let $(x,1)$ for a vector  $x\in \RK$
denote the vector $y\in \RKw$ such that $y(j)=x(j), j\in [1,K]$
and $y(K+1)=1$. 

Choose an integer $p_{\max}\geq 1$ as the dimension of a search space $\R^{p_{\max}}$,
a vector $s^*\in  \R^{p_{\max}}$ as the center of  $\R^{p_{\max}}$
and a vector $\delta\in  \R_+^{p_{\max}}$ with 
  $\delta(p)>0, p\in[1,p_{\max}]$ as the width of a grid   in the space $\R^{p_{\max}}$.
A {\em grid} is defined to be  an integer vector  $z\in \Z^{p_{\max}}$
for which we define a subspace 
 $S(z)\subseteq  \R^{p_{\max}}$  to be 
\[  S(z)\triangleq \{s\in  \R^{p_{\max}}  \mid  
           s^*(p)+(z(p)-0.5)\delta(p)  \leq   s(p)\leq s^*(p)+(z(p)+0.5)\delta(p), p\in[1,p_{\max}] \} .\]
We call a grid  $z\in \Z^{p_{\max}}$ with $z(p)=0, p\in[1,p_{\max}]$ the {\em center grid}.
A neighbor  $N(r)$ of the center  with a radius vector $r\in \Z_+^{p_{\max}}$ is defined
to be a set  of grids such that 
\[ N(r)\triangleq 
\{  z\in \Z^{p_{\max}} \mid -r(p) \leq z(p)\leq r(p), p\in[1,p_{\max}] \}. \]
% The {\em   distance} $\mathrm{dst}(z)$ of a grid $z\in Z^{p_{\max}}$
% to the origin $s^*$ is defined to be 
% $\mathrm{dst}(z)\triangleq \sum_{p:z(p)\geq 0}|z(p)| +\sum_{p:z(p)< 0}|z(p)+1|$,
% where there are $2^{p_{\max}}$ grids $z$ with $\mathrm{dst}(z)=0$.
%
Let us introduce a partial order $\preceq$ over the set  $\Z^{p_{\max}}$ of grids.
For two grids $z,z'\in \Z^{p_{\max}}$, $z'\preceq z $ if
$0\leq z'(p)\leq z(p)$ or $0\geq z'(p)\geq z(p)$ 
for each $p\in[1,p_{\max}]$, where we let $z'\prec z$ mean that
$z'\preceq z$ and $z'\neq z$.

% For a grid  $z\in \Z^{p_{\max}}$, let $\mathrm{HC}(z)$ denote
% the minimal hypercube in the space $\Z^{p_{\max}}$ 
% that contains $z$ and the origin $0\in \Z^{p_{\max}}$;
% i.e., $\mathrm{HC}(z) \triangleq \{z' \in \Z^{p_{\max}} \mid
% 0\leq z'(p)\leq z(p) \mbox{ or } 0\geq z'(p)\geq z(p) 
% \mbox{ for each }p\in[1,p_{\max}]  \}$.
 
We introduce a linear function $\theta_p: \RK\to \R$ for each $p\in[1,p_{\max}]$,
called a {\em projection function}  such that 
\[ \theta_p(x) \triangleq w_p\cdot (x,1)=\sum_{j\in [1,K]}w_p(j)x(j) + w_p(K+1) \]
by choosing a vector $w_p\in \RKw$. 
For a notational convenience, let $\theta(x), x\in \RK$ denote the vector 
$(\theta_1(x), \theta_2(x), \ldots, \theta_{p_{\max}}(x)) \in \R^{p_{\max}}$. 

\bigskip 
In the framework of inferring chemical graphs, 
we formulate an MILP $\mathcal{M}(g, x,y;\mathcal{C}_1,\mathcal{C}_2)$
that consists of   two MILPs 
$\mathcal{M}(x,y;\mathcal{C}_1)$ and $\mathcal{M}(g,x;\mathcal{C}_2)$,
where the former simulates 
the computation process of  a prediction function $\eta$ for a chemical property $\pi$ and 
the latter  simulates the computation process of   a feature function $f$
and describes 
construction of a chemical graph that satisfies a given topological specification $\sigma$.
By solving the MILP for lower and upper bounds on
a  target value  of $\pi$, we obtain a desired chemical graph $\C^\dagger$ 
when the MILP instance is feasible (or we detect that there is no such chemical graph
when the instance is infeasible).

We design a procedure for finding other solutions of the MILP by
searching the neighbor of the output chemical graph  $\C^\dagger$.
Suppose that we have chosen
 a dimension $p_{\max}$, 
 a width vector  $\delta\in \R_+^{p_{\max}}$ with 
  $\delta(p)>0, p\in[1,p_{\max}]$, 
 a radius vector $r\in \Z_+^{p_{\max}}$ and 
 projection functions $\theta_p, p\in[1,p_{\max}]$. 

\begin{enumerate}
\item[1.] 
Let  $\underline{y}^*$ and $\overline{y}^*$ be lower and upper bounds on
the value of property $\pi$ of a chemical graph to be inferred.
We first solve $\mathcal{M}(g, x, y;\mathcal{C}_1,\mathcal{C}_2)$ with
 the input values $(\underline{y}^*,\overline{y}^*)$ 
 to find a desired chemical graph  $\C^\dagger$.
 (When the MILP instance is infeasible, we halt.)
Let $x^*:=f(\C^\dagger)\in \RK$. 

\item[2.]  Set the center of the space $\R^{p_{\max}}$ to be
$s^*:=\theta(x^*)$. 
For each grid $z \in N(r)$, we solve
an MILP $\mathcal{M}(g, x,y;\mathcal{C}_1,\mathcal{C}_2)$ 
with an additional linear constraint $\theta(x)\in S(z)$ 
for $(\underline{y}^*,\overline{y}^*)$,
where we call a grid $z$ {\em feasible} (resp., {\em infeasible}) 
if the augmented MILP instance is feasible  (resp., infeasible). 
 
\item[3.]
For each feasible grid $z \in N(r)$, output a feasible solution $\C^\dagger_z$
of  the augmented MILP instance.
We check the feasibility of grids  $z \in N(r)$ in a non-decreasing order of
$\max_{p\in [1,p_{\max}]}|z(p)|$, and discard any grid $z \in N(r)$
without testing the feasibility of $z$ if we find an infeasible grid $z'\in N(r)$
with $z'\prec z$.
Note that $\C^\dagger_z$ and $\C^\dagger_{z'}$ are not isomorphic unless
$\theta(f(\C^\dagger_z))$ and $\theta(f(\C^\dagger_{z'}))$ happen to belong to
the common boundary $S(z) \cap S(z')$.
\end{enumerate}

In the above method, we can choose arbitrarily many grids in the space of  $\R^{p_{\max}}$
around the center $s^*$ by choosing small   $\delta(p)$ and large  $r(p),  p\in[1,p_{\max}]$,
where  each $\delta(p)$ needs to be large enough to avoid a possible numerical error. 

We can also choose arbitrary linear functions as projection functions  $\theta_p,  p\in[1,p_{\max}]$.
When we have constructed a prediction function $\eta_\tau$ as a linear function
with linear regression for several chemical properties $\tau$ other than the current
target property $\pi$, we can use such functions as projection functions.
For example, if  a linear prediction function $\eta_\tau$ is available 
for a chemical property $\tau$ such as 
%octanol/water partition coefficient ({\sc Kow}), 
solubility ({\sc Sl}) and lipophilicity ({\sc Lp}), 
then we can infer chemical graphs $\C^\dagger_z$ with slightly different values
of these properties  $\tau_1= $ {\sc Sl}   and $\tau_2=$ {\sc Lp} 
by setting $\theta_p:=\eta_{\tau_p}, p=1,2$. 

%\input how_to_implement.tex

%\clearpage 

\section{Results}\label{sec:experiment}%%%%%%%%

We implemented our method of Stages~1 to 5 
for inferring chemical graphs under a given topological specification and
conducted experiments  to evaluate the computational efficiency. 
We executed the experiments on a PC with 
 Processor:  Core i7-9700 (3.0GHz; 4.7 GHz at the maximum) and 
Memory: 16 GB RAM DDR4. 
% We used  ChemDoodle  version 10.2.0  for constructing  2D drawings of 
 % chemical graphs.  
To construct an ANN, 
we used {\tt scikit-learn} version 0.23.2  with Python 3.8.5, 
MLPRegressor and ReLU activation function. 

\medskip \noindent
{\bf Results on Phase~1.  }
We implemented Stages~1, 2 and 3 in Phase~1 as follows.

We have conducted experiments of Lasso linear regression and
 for 37 chemical properties of monomers
 (resp., ten chemical properties of polymers)  using the same feature function in this paper 
 and  we found that 
 the test  coefficient of determination ${\rm R}^2$ 
 exceeds 0.927 for the following 11 properties of monomers:  
 octanol/water partition coefficient ({\sc Kow}), 
heat of combustion ({\sc Hc}), 
vapor density ({\sc Vd}),  
electron density on the most positive atom ({\sc EDPA}), 
heat of atomization ({\sc Ha}), 
heat of formation ({\sc Hf}), 
internal energy at 0K ({\sc U0}),  
isotropic polarizability ({\sc Alpha}), 
heat capacity at 298.15K ({\sc Cv}),  
isobaric heat capacities in liquid phase ({\sc IhcLiq}) and 
isobaric heat capacities in solid phase ({\sc IhcSol}) 
(see \cite{ZAHZNA21} for the details)
and
that the test  coefficient of determination ${\rm R}^2$ 
 exceeds 0.9  for the following five properties of polymers:  
experimental amorphous density~({\sc AmD}), 
heat capacity liquid ({\sc HcL}), 
heat capacity solid ({\sc HcS}), 
mol volume ({\sc MlV}) and 
glass transition~({\sc Tg})
(see \cite{ICZAHZNA21}  for the details).
We excluded the above properties in our experiment of 
constructing prediction functions with ANNs.

We have conducted experiments of ANNs 
 for the rest of 26 chemical properties of monomers
 (resp., five chemical properties of polymers) among which 
we report the following 12 properties of monomers
(resp., two properties of polymers) to which the test  
 coefficient of determination ${\rm R}^2$ 
by ANNs is better than that by Lasso linear regression:
biological half life ({\sc BHL}), %[1]
boiling point ({\sc Bp}),  %[4]
critical pressure ({\sc Cp}), %[1]
dissociation constants  ({\sc Dc}), %[1]
flash point   ({\sc Fp}), %[1]
Kovats retention index   ({\sc Kov}), %[2] 
lipophilicity ({\sc Lp}), %[10]
energy of lowest unoccupied molecular orbital ({\sc Lumo}), %[6] 
optical rotation ({\sc OptR}), %[1]
solubility ({\sc Sl}), %[11]
surface tension ({\sc SfT}) and  %[7]
viscosity ({\sc Vis})  %[8]
(resp.,  
characteristic ratio ({\sc ChaR}) and 
refractive index ({\sc RfId})).
We explain the data set and the results for these 14 properties 
in detail below.

We used data sets of monomers  provided  by HSDB from PubChem~\cite{pubchem} %[1]
 for {\sc BHL},  {\sc Cp},  {\sc Dc},   {\sc Fp}   and  {\sc OptR}, %[1]
 M.~Jalali-Heravi and M.~Fatemi~\cite{JF01} for {\sc Kov},  %  [2]
   Roy and Saha~\cite{RS03} for {\sc Bp}, %[4] 
 MoleculeNet~\cite{moleculenet} for  {\sc Lumo},  %[6]
Goussardet al.~\cite{GFPDNA17}  for {\sc SfT},  %[7] 
 Goussard et al.~\cite{GFPDNA20} for   {\sc Vis} and %[8]  
 Figshare~\cite{figshare}  for {\sc Lp}. %[10]
Property  {\sc  Lumo} has the original data set $D^*$
with  more than 130,000 compounds, and 
we used  a set $D_\pi$ of 1,000 graphs randomly selected from $D^*$  
as a   data set of   property  {\sc  Lumo}
in this experiment. 
 
We used data sets of polymers  provided  by   
Bicerano~\cite{Bicerano}, %[12] 
where we did not include any polymer whose chemical formula
could not be found by its name in the book. 
For property {\sc ChaR} (resp.,  {\sc RfId}), we remove the following   polymer 
as an outlier  from the original data set: \\
 ethyleneTerephthalate,                              % value:3.7
 oxy(2-methyl-6-phenyl-1$\underline{~}$4-phenylene)  and % value:4.1
 N-vinylCarbazole                                      % value:15.9
%Polyacrylonitrile value: 0.52  and 
(resp.,  2-decyl-1$\underline{~}$4-butadiene). %value: 0.4899  

\medskip \noindent
{\bf Stage~1.  }
We set  a graph class   $ \mathcal{G}$   to be
the set of all chemical graphs with any graph structure, 
and set a branch-parameter ${\rho}$ to be 2. 

For each of the properties,  
 we first select a set $\Lambda$ of chemical elements 
 and then collect  a  data set  $D_{\pi}$ on chemical graphs
 over the set $\Lambda$ of chemical elements.  
 To construct the data set $D_{\pi}$,
  we eliminated  chemical compounds that do not satisfy 
  one of the following: the graph is connected,
  the number of carbon atoms is at least four,
  and   the number of non-hydrogen neighbors of each atom is
  at most 4.

\medskip \noindent
{\bf Stage~2.  }
We used  the  new  feature function  defined 
in our chemical model without suppressing hydrogen 
(see Appendix~\ref{sec:descriptor} for the detail).
We normalize the range of each descriptor and
 the range $\{t\in \R \mid \underline{a}\leq t\leq \overline{a}\}$ 
 of property values   $a(\Co), \Co\in D_\pi$.

 Table~\ref{table:phase1a}    shows
  the size and range of data sets   that 
 we prepared for each chemical property in Stages~1 and 2,
 where  we denote the following:  
\begin{enumerate}[nosep,  leftmargin=*]
\item[-] 
  $\Lambda$: the set of elements used in the data set $D_{\pi}$; 
  $\Lambda$ is one of the following nine sets: 
  $\Lambda_1=\{\ttH,\ttC,\ttO \}$; 
   $\Lambda_2=\{\ttH,\ttC,\ttO, \ttN \}$;
  % $\Lambda_3=\{\ttH,\ttC,\ttO, \ttCl \}$;
   $\Lambda_3=\{\ttH,\ttC,\ttO, \ttSi_{(4)} \}$;  
  $\Lambda_4=\{\ttH,\ttC,\ttO, \ttN,\ttS_{(2)},\ttF \}$;
   $\Lambda_5=\{\ttH,\ttC,\ttO, \ttN,  \ttCl, \ttPb \}$;
   $\Lambda_6=\{\ttH,\ttC,\ttO, \ttN,\ttS_{(2)},\ttS_{(6)},\ttCl \}$;    
   $\Lambda_7=\{\ttH,\ttC,\ttO, \ttN,\ttS_{(2)},\ttS_{(4)},\ttS_{(6)},\ttCl \}$;  
    $\Lambda_{8}=\{\ttH, \ttC_{(2)},\ttC_{(3)},\ttC_{(4)},\ttC_{(5)},\ttO,
   \ttN_{(1)}, \ttN_{(2)},$ $ \ttN_{(3)}, \ttF \}$;
  $\Lambda_{9}=\{\ttH,\ttC, \ttO_{(1)}, \ttO_{(2)}, \ttN \}$; % C4 O1 O2 N3 H1
 $\Lambda_{10}=\{\ttH,\ttC,\ttO, \ttN,\ttSi_{(4)},\ttCl,\ttBr \}$;  % C4 O2 N3 Si4 Cl1 Br1 H1
  $\Lambda_{11}=\{\ttH,\ttC,\ttO_{(1)}, \ttO_{(2)},\ttN,\ttSi_{(4)},\ttCl,\ttF \}$; 
   and % C4 O1 O2 N3 Si4 Cl1 F1 H1
  $\Lambda_{12}=\{\ttH,\ttC,\ttO_{(1)}, \ttO_{(2)}, \ttN,\ttSi_{(4)},\ttCl,\ttF,
                            \ttS_{(2)}, \ttS_{(6)},\ttBr \}$, % C4 O1 O2 N3 Si4 Cl1 F1 S2 S6 Br1 H1
 where ${\tt a}_{(i)}$ for a chemical element ${\tt a}$ and an integer $i\geq 1$ 
 means that  a chemical element ${\tt a}$ with valence~$i$. 

\item[-] 
 $|D_{\pi}|$:  the size of data set $D_{\pi}$ over $\Lambda$
  for the property $\pi$.
  
\item[-]   $ \underline{n},~\overline{n} $:  
  the minimum and maximum  values of the number 
  $n(\Co)$ of non-hydrogen atoms in 
  the   compounds $\Co$ in $D_{\pi}$.
\item[-] $ \underline{a},~\overline{a} $:  the minimum and maximum values
of $a(\Co)$ for $\pi$ over   the   compounds $\Co$ in  $D_{\pi}$.
\item[-]    $|\Gamma|$: 
the number of different edge-configurations
of interior-edges over the compounds in~$D_{\pi}$. 
\item[-]  $|\mathcal{F}|$: the number of non-isomorphic chemical rooted trees
 in the set of all 2-fringe-trees in  the   compounds in $D_{\pi}$.
 
\item[-] $K$: the number of descriptors in the original feature vector $f(\Co)$.  
\end{enumerate}

\medskip \noindent
{\bf Stage~3.  }
For each chemical property $\pi$, we conducted a preliminary experiment
to choose the following:  
 a subset $S_\pi$ of the original set of $K$ descriptors; 
  an architecture $A_\pi$ with at most five hidden layers; 
   a nonnegative real $\rstp_\pi\leq 1$; and
  an integer $\itstp_\pi$, where we will use 
  $\rstp_\pi$ and $\itstp_\pi$ as parameters to execute an early stopping 
  in constructing a prediction function with a training data set. 
 Let $f_\pi$ denote the feature vector that consists of the descriptors in the set  $S_\pi$.
 
For  each  property $\pi$, we conducted ten 5-fold cross-validations.
In a 5-fold cross-validation, 
  we construct  five prediction functions  $\eta^{(k)}, k\in[1,5]$ as follows.
 Partition data set $D_{\pi}$ 
 into five subsets $D_{\pi}^{(k)}$, $k\in[1,5]$ randomly.
 For each $k\in[1,5]$, use the  set $\Dtrain:=D_{\pi}\setminus D_{\pi}^{(k)}$ as a training set
 and  construct an ANN on the selected architecture $A_\pi$
 with the feature vector  $f_\pi$ by the MLPRegressor of {\tt scikit-learn},
 where we stop updating weights/biases on $A_\pi$ during an execution of the iterative
 algorithm when  the coefficient of determination 
 $\mathrm{R}^2(\eta,\Dtrain)$ of the prediction function
 $\eta$ by the current weights/biases  exceeds  $\rstp_\pi$
 (where we terminate the execution when the number of iterations exceeds
 $1.5\times \itstp_\pi$ even if 
 $\mathrm{R}^2(\eta,\Dtrain)$ does not reach  $\rstp_\pi$).
Set $\eta^{(k)}$ to be the  prediction function
 $\eta$ by the resulting weights/biases on $A_\pi$.
We evaluate the performance of the prediction function  $\eta^{(k)}$
with the coefficient  $\mathrm{R}^2(\eta^{(k)},\Dtest)$ 
  of determination  for the test set $\Dtest:=D_{\pi}^{(k)}$.   
The running time per trial in a cross-validation was at most 8.4 seconds.  
  
\begin{table}[h!]\caption{Results of Stages 1 and 2 in Phase 1.} 
  \begin{center}
    \begin{tabular}{@{} c c r c  c  r r r   @{}}\hline
      $\pi$ & $\Lambda$  &  $|D_{\pi}|$  &  $ \underline{n},~\overline{n} $ &   $\underline{a},~\overline{a}$ &
   $|\Gamma|$   &  $|\mathcal{F}|$ &   $K$~ \\ \hline
      {\sc BHL} & $\Lambda_2$  & 300  &  5,\,36  &  0.03,\,732.99   & 20   & 70  & 120  \\
      {\sc BHL} & $\Lambda_7$  & 514  &  5,\,36  &  0.03,\,732.99   & 26  & 101  & 166    \\
      {\sc Bp} & $\Lambda_2$  & 370  &  4,\,67  &  -11.7,\,470.0   & 22   & 130  & 184      \\
      {\sc Bp} & $\Lambda_6$  & 444  &  4,\,67  &  -11.7,\,470.0   & 26   & 163 & 230     \\
      {\sc Cp} & $\Lambda_2$  & 125 &  4,\,63  &  $4.7\!\times\!10^{-6}$,\,5.52   & 8  & 75  & 112     \\
      {\sc Cp} & $\Lambda_5$  & 131  &  4,\,63  &  $4.7\!\times\! 10^{-6}$,\,5.52  & 8   & 79  & 119       \\ 
      {\sc Dc} & $\Lambda_2$  & 141  &  5,\,44  &  0.5,\,17.11   & 20   & 62  & 111     \\ 
      {\sc Dc} & $\Lambda_6$  & 161  &  5,\,44  &  0.5,\,17.11   & 25   & 69  & 130     \\ 
      {\sc Fp} & $\Lambda_2$  & 368  &  4,\,67  &  -82.99,\,300.0   & 20   & 131  & 183      \\ 
      {\sc Fp} & $\Lambda_6$  & 424  &  4,\,67  &  -82.99,\,300.0   & 25   & 161  & 229       \\ 
      {\sc Kov} & $\Lambda_1$  & 52  &  11,\,16  &  1422.0,\,1919.0   & 9   & 33  & 64      \\ 
      {\sc Lp} & $\Lambda_2$  & 615  &  6,\,60  &  -3.62,\,6.84   & 32   & 116  & 186      \\ 
      {\sc Lp} & $\Lambda_7$  & 936  &  6,\,74  &  -3.62,\,6.84  & 44   & 136  & 231     \\ 
      {\sc Lumo} & $\Lambda_8$  & 977  &  6,\,9  &  -0.1144,\,0.1026   & 59   & 190 & 297   \\ 
      {\sc OptR} & $\Lambda_2$  & 147  & 5,\,44  &  -117.0,\,165.0   & 21   & 55  & 107      \\ 
      {\sc OptR} & $\Lambda_4$  & 157  &  5,\,69  &  -117.0,\,165.0   & 25   & 62  & 123     \\ 
      {\sc Sl} & $\Lambda_2$  & 673  &  4,\,55  &  -9.332,\,1.11   & 27   & 154  & 217     \\ 
      {\sc Sl} & $\Lambda_7$  & 915 &  4,\,55  &  -11.6,\,1.11  & 42   & 207  & 300      \\ 
      {\sc SfT} & $\Lambda_3$  & 247  &  5,\,33  & 12.3,\,45.1  & 11   & 91  & 128      \\ 
      {\sc Vis} & $\Lambda_3$  & 282  &  5,\,36  &  -0.64,\,1.63   & 12   & 88  & 126      \\ 
      {\sc ChaR} & $\Lambda_2$  & 27  &  4,\,18  &  5.5,\,13.2  & 22   & 17  & 67 \\ % C4 O2 N3 Cl1 H1
      {\sc ChaR} & $\Lambda_{10}$  & 32  & 4,\,18  &  5.5,\,13.2   & 26   & 21  & 82  \\    % C4 O2 N3 Si4 Cl1 Br1 H1
      {\sc RfId} & $\Lambda_{9}$  & 91  &  4,\,29  &  1.339,\,1.683   & 26   & 35  & 96 \\ % C4 O1 O2 N3 H1
      {\sc RfId} & $\Lambda_{11}$  & 124 & 4,\,29  &  1.339,\,1.683  & 32 & 50  & 124 \\ % C4 O1 O2 N3 Si4 Cl1 F1 H1
      {\sc RfId} & $\Lambda_{12}$  & 134  &  4,\,29  &  1.339,\,1.71   & 38   & 56  & 144  \\ % C4 O1 O2 N3 Si4 Cl1 F1 S2 S6 Br1 H1
      \hline
  \end{tabular}\end{center}\label{table:phase1a}
\end{table} 

\begin{table}[h!]\caption{Results of Stage 3 in Phase 1.} 
  \begin{center}
    \begin{tabular}{@{} c  r    c c  c c  @{}}\hline
      $\pi$    &  $|D_{\pi}|$  &      $A_\pi$ &  $\rstp_\pi$ &  ANN $\mathrm{R}^2$  &  LLR $\mathrm{R}^2$ \\ \hline
      {\sc BHL}    & 300      & $(108,64,64,64,64,64,1)$ & 0.86  & 0.630 &  0.364  \\
      {\sc BHL}    & 514 & $(46,10,8,1)$ & 0.71 & 0.622 &  0.483  \\
      {\sc Bp}   & 370    & $(71,28,22,17,13,10,1)$ & 0.93 & 0.765 & 0.599   \\
      {\sc Bp}   & 444    & $(225,135,135,135,135,135,1)$ & 0.98 & 0.720 &  0.663  \\
      {\sc Cp}    & 125    & $(19,10,10,10,1)$ & 0.66 & 0.694 &  0.445  \\
      {\sc Cp}    & 131    & $(19,10,10,10,1)$ & 0.66 & 0.727 & 0.556   \\ 
      {\sc Dc}   & 141    & $(53,10,8,1)$ & 0.93 & 0.651&  0.489  \\ 
      {\sc Dc}    & 161    & $(109,87,87,87,87,1)$ & 0.94 & 0.622 &  0.574  \\ %NG
      {\sc Fp}    & 368      & $(30,10,10,10,10,10,1)$ & 0.88 & 0.746 &  0.589  \\ 
      {\sc Fp}   & 424     & $(42,16,16,16,16,1)$ & 0.90 & 0.733 & 0.571   \\ 
      {\sc Kov}    & 52      & $(28,10,10,1)$ & 0.92 & 0.727 & 0.677   \\ 
      {\sc Lp}   & 615      & $(186,74,74,74,74,74,1)$ & 0.98 & 0.867 &  0.856    \\ %NG
      {\sc Lp}   & 936    & $(197,157,157,157,157,157,1)$ & 0.81 & 0.859 &  0.840   \\ %NG
      {\sc Lumo}    & 977   & $(241,192,192,192,192,192,1)$ & 0.99 & 0.860 &   0.841   \\ %NG
      {\sc OptR}    & 147    & $(107,64,51,40,32,1)$ & 0.97 & 0.919 &  0.823  \\ 
      {\sc OptR}    & 157      & $(114,20,10,1)$ & 0.96 & 0.894 &  0.825  \\ 
      {\sc Sl}    & 673   & $(205,10,5,1)$ & 0.94 & 0.819 &  0.772  \\ 
      {\sc Sl}    & 915    & $(126,25,25,25,1)$ & 0.95 & 0.822 &  0.808  \\ 
      {\sc SfT}     & 247     & $(19,15,12,9,1)$ & 0.91 & 0.834 &  0.804  \\ 
      {\sc Vis}    & 282     & $(19,11,8,6,5,1)$ & 0.97 & 0.929 &  0.893  \\ 
      {\sc ChaR}    & 27   & $(38,22,22,22,22,22,1)$ & 0.98 & 0.641 &  0.431 \\ % C4 O2 N3 Cl1 H1
      {\sc ChaR}    & 32   & $(38,15,15,15,1)$ & 0.90 & 0.622 &  0.235  \\ % C4 O2 N3 Si4 Cl1 Br1 H1
      {\sc RfId}    & 91  & $(31,18,14,11,8,6,1)$ & 0.95 & 0.871&  0.852  \\ % C4 O1 O2 N3 H1
      {\sc RfId}    & 124   & $(60,48,38,30,24,1)$ & 0.90 & 0.891 &  0.832  \\ % C4 O1 O2 N3 Si4 Cl1 F1 H1
      {\sc RfId}    & 134  & $(53,42,33,26,1)$ & 0.94 & 0.866 &  0.832  \\ % C4 O1 O2 N3 Si4 Cl1 F1 S2 S6 Br1 H1
      \hline
  \end{tabular}\end{center}\label{table:phase1b}
\end{table}

 Table~\ref{table:phase1b}   shows the results on Stage~3,
 where  we denote the following:     
\begin{enumerate}[nosep,  leftmargin=*]  
\item[-]  
$A_\pi$: an architecture $A_\pi$  used to construct a prediction function for property $\pi$,
 where  $(K',p_1,p_2,$ $\ldots,p_\ell, 1)$ means an architecture
 with   an input layer with $K'$ nodes, 
 $\ell$ hidden layers  with $p_i, i\in[1,\ell]$ nodes 
and an output layer with a single node, where $K'$ is
the size $|S_\pi|$  of the set $S_\pi$ of selected  descriptors from the original set of $K$ descriptors.
  
\item[-]
$\rstp_\pi$:  
  a nonnegative real with $0\leq \rstp_\pi\leq 1$ by which we execute an early stopping 
  in constructing a prediction function with a training data set. 
  
\item[-]
ANN $\mathrm{R}^2$: the median of test $\mathrm{R}^2$ over all 50 trials
  in ten 5-fold cross-validations for prediction functions constructed
  with ANNs in this paper.   
  
\item[-]
LLR $\mathrm{R}^2$: the median of test $\mathrm{R}^2$ over all 50 trials
  in ten 5-fold cross-validations for prediction functions constructed
  with Lasso linear regression~\cite{ZAHZNA21}.  
\end{enumerate}

%\clearpage 

\medskip \noindent
{\bf Results on Phase~2.  }
To execute  Stages~4  and 5 in Phase~2, 
we used a set of seven instances
$I_{\mathrm{a}}$, $I_{\mathrm{b}}^i, i\in[1,4]$, $I_{\mathrm{c}}$
 and $I_{\mathrm{d}}$ based on the seed graphs prepared by Zhu et~al.~\cite{ZAHZNA21}. 
We here present their seed graphs $\GC$ 
(see Appendix~\ref{sec:specification} for the details of $I_{\mathrm{a}}$
and Appendix~\ref{sec:test_instances} for the details of 
$I_{\mathrm{b}}^i, i\in[1,4]$, $I_{\mathrm{c}}$  and $I_{\mathrm{d}}$).

\begin{figure}[h!] \begin{center}
\includegraphics[width=.85\columnwidth]{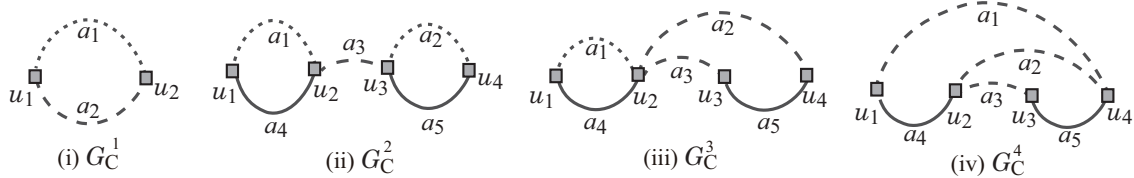}
\end{center} \caption{%An illustration of seed graphs:
%  for inferring cyclic graphs with rank at most 2: 
(i)  Seed graph   $\GC^1$ for $I_{\mathrm{b}}^1$ and  $I_{\mathrm{d}}$;
(ii) Seed graph $\GC^2$   for $I_{\mathrm{b}}^2$; 
(iii) Seed graph  $\GC^3$   for $I_{\mathrm{b}}^3$; 
(iv)  Seed graph $\GC^4$  for $I_{\mathrm{b}}^4$. }
\label{fig:specification_example_polymer}
\end{figure}

The seed graph  $\GC$ of  $I_{\mathrm{a}}$ is given
 by the graph in Figure~\ref{fig:specification_example_1}(a).
The seed graph $\GC^1$ of  $I_{\mathrm{b}}^1$
(resp., $\GC^i, i=2,3,4$ of $I_{\mathrm{b}}^i,  i=2,3,4$) is illustrated
 in Figure~\ref{fig:specification_example_polymer}.
 
\begin{figure}[!htb]
\begin{center} 
 \includegraphics[width=.69\columnwidth]{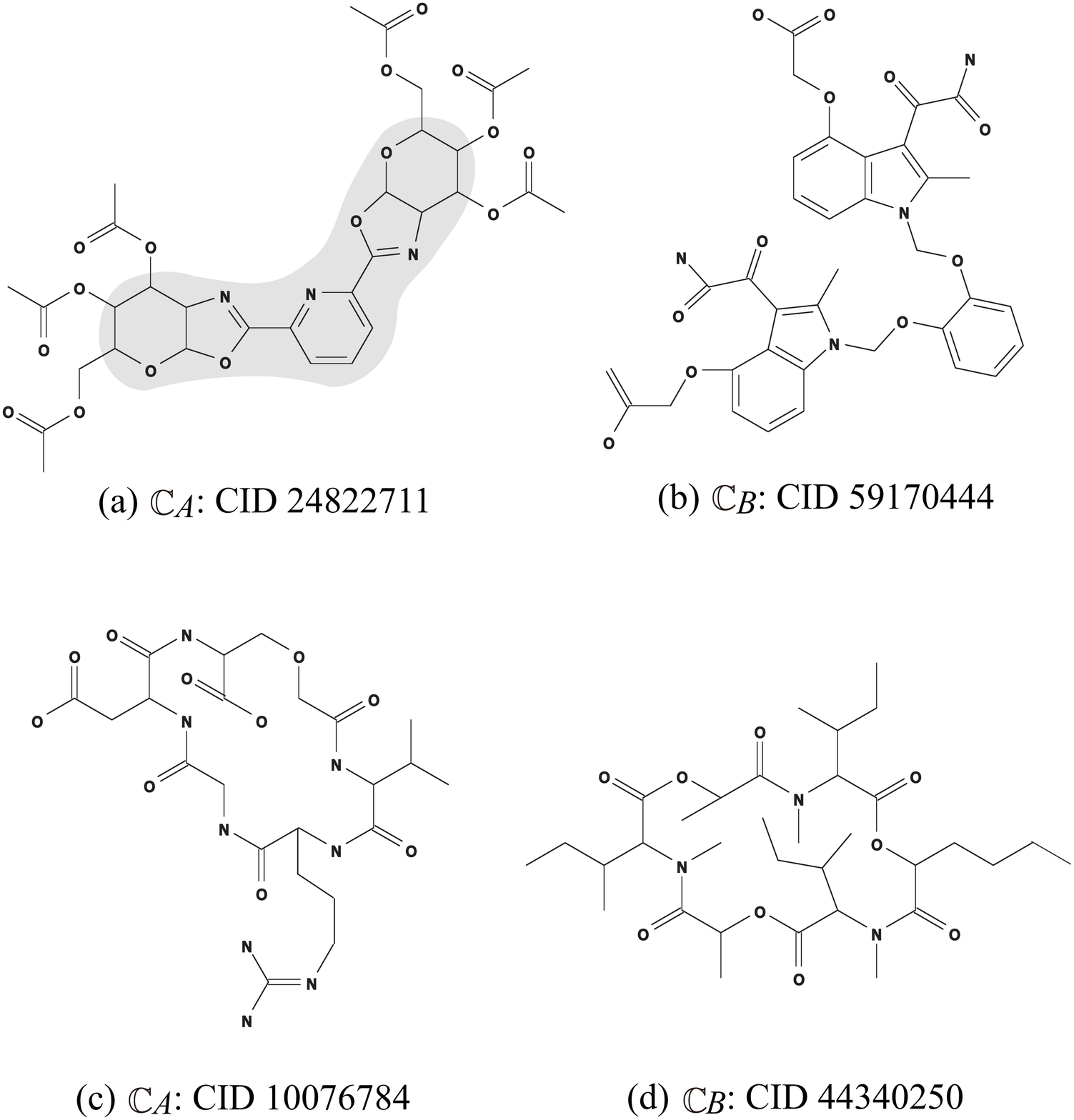}
\end{center}
\caption{An illustration of  chemical compounds 
  for instances  $I_{\rm c}$  and  $I_{\rm d}$: 
(a) $\Co_A$: CID~24822711;
(b)  $\Co_B$: CID~59170444; 
(c) $\Co_A$: CID~10076784;
(d)  $\Co_B$: CID~44340250,
where hydrogens are omitted. 
}
\label{fig:instance_I_c_I_d}  
\end{figure}

Instance  $I_{\mathrm{c}}$ has been introduced 
in order to infer a chemical graph $\Co^\dagger$ such that
the core of $\Co^\dagger$ is equal to the core of 
chemical graph $\Co_A$: CID~24822711 in Figure~\ref{fig:instance_I_c_I_d}(a)
and 
the frequency of each edge-configuration in the non-core of $\Co^\dagger$
is equal to that of chemical graph  $\Co_B$:  CID~59170444
 in  Figure~\ref{fig:instance_I_c_I_d}(b).
This means that the seed graph  $\GC$ of   $I_{\mathrm{c}}$
 is the core of $\Co_A$
which is indicated by a shaded area in  Figure~\ref{fig:instance_I_c_I_d}(a). 

Instance  $I_{\mathrm{d}}$ has been introduced 
in order to   infer a chemical monocyclic graph $\Co^\dagger$ such that
the frequency vector of  edge-configurations in  $\Co^\dagger$
is a vector obtained by merging those of chemical graphs 
$\Co_A$: CID~10076784   and $\Co_B$: CID~44340250 
in   Figure~\ref{fig:instance_I_c_I_d}(c) and (d), respectively.  
The seed graph  $\GC$ of    $I_{\mathrm{d}}$  is given by  $\GC^1$  
 in Figure~\ref{fig:specification_example_polymer}(i).

\medskip \noindent
{\bf Stage~4.  } 
We executed Stage~4 for three properties $\pi\in \{${\sc FP,  OptR,  SfT}$\}$.  

 For the MILP formulation  $\mathcal{M}(x,y;\mathcal{C}_1)$,
we use the prediction function  
 that attained the median  test $\mathrm{R}^2$ in Table~\ref{table:phase1b}.
 To solve an MILP   in Stage~4, we used % CPLEX~\cite{cplex}.
{\tt  CPLEX} version 12.10.
Tables~\ref{table:stages_4_5_Fp} and \ref{table:stages_4_5_SfT}  show
   the computational results of the experiment
in Stage~4 for the five properties, 
 where we denote the following:
\begin{enumerate} [nosep,  leftmargin=*]
  
\item[-]  
$n_\LB$: a lower bound on the number of non-hydrogen atoms 
in  a chemical graph $\Co$ to be inferred; 

\item[-]  
  $ \underline{y}^*,~\overline{y}^* $:  
 lower and upper bounds $\underline{y}^*, \overline{y}^*\in \R$ 
  on the value $a(\Co)$ of a chemical graph $\Co$ to be inferred; 
  
\item[-]  
 $\#$v (resp.,  $\#$c): 
 the number  of variables (resp., constraints)  in the MILP  in Stage~4;  
  
\item[-]   
 I-time: the   time (sec.) to solve the MILP  in Stage~4; 

\item[-]  
    $n$:  the number  $n(\Co^\dagger)$  of  non-hydrogen atoms
     in the chemical graph $\Co^\dagger$   inferred in Stage~4;   
     
\item[-]  
  $\nint$:  the number  $\nint(\Co^\dagger)$ of interior-vertices in
  the chemical graph $\Co^\dagger$   inferred in Stage~4; and 
      
\item[-]  
$\eta$: the predicted property value 
$\eta(f(\Co^\dagger))$ of the chemical graph $\Co^\dagger$ inferred 
in Stage~4.
\end{enumerate}

\begin{figure}[!htb]
\begin{center} 
\includegraphics[width=.98\columnwidth]{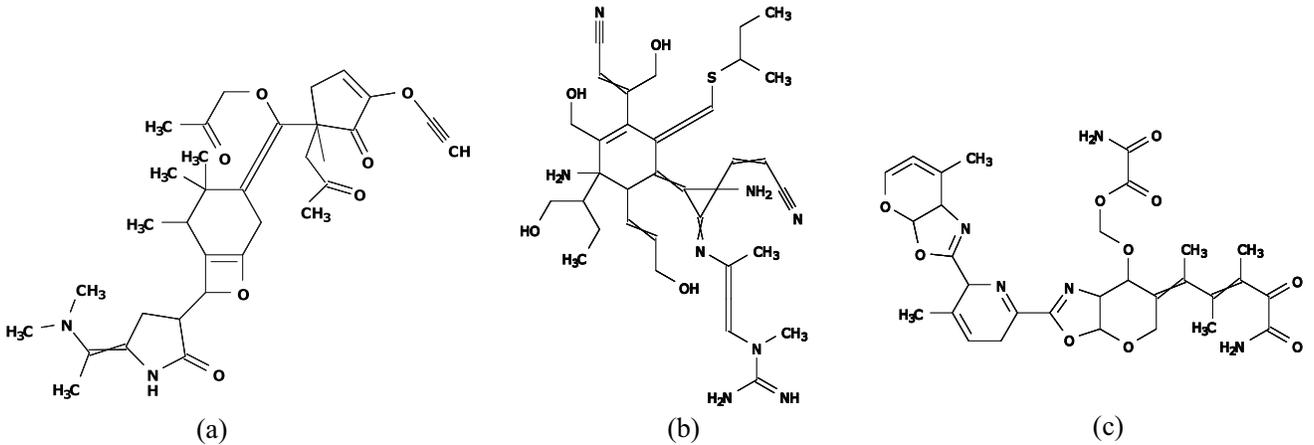}
\end{center}
\caption{ 
(a)  $\Co^\dagger$ with   $\eta(f(\Co^\dagger))= 132.827$ inferred
  from   $I_{\mathrm{a}}$ with $(\underline{y}^*, \overline{y}^*) =(137,\,140 )$  of  {\sc Fp};  
(b)  $\Co^\dagger$ with   $\eta(f(\Co^\dagger))=51.452$ inferred
 from   $I_{\mathrm{b}}^2$ with $(\underline{y}^*, \overline{y}^*) =(50, 52)$  of  {\sc OptR}; and   
(c)  $\Co^\dagger$ with  
  $\eta(f(\Co^\dagger))=44.557$ inferred
  from   $I_{\mathrm{c}}$ with $(\underline{y}^*, \overline{y}^*) =(44, 45)$  of  {\sc SfT}.   
}
\label{fig:MILP_solutions}  
\end{figure}

\begin{table}[h!]\caption{ Results of Stages~4 and 5 for  {\sc Fp}.}  
 \begin{center}
 \begin{tabular}{@{}  c  c   c  r r r r r c  r r r     @{}}\hline                
 inst. & $n_\LB$ &  $ \underline{y}^*,~\overline{y}^* $ & $\#$v~  &  $\#$c~~   &  
   {\small I-time}\!\! & $n$~  &  \!\!$\nint$  &  $\eta $ \!\! & 
                 {\small  D-time} &  {\small $\Co$-LB} &  {\small $\#\Co$}    \\ \hline
   $I_{\mathrm{a}}$ & 30 &  130,\,133 &10877 &10604& 13.7 &42 &25 &132.827 &0.0703& 1 &1   \\%
  $I_{\mathrm{b}}^1$ & 10 & 110,\,113 &10773 & 8255 &1.99 &10 & 7 &112.428 &0.0234 & 2 &2  \\%
  $I_{\mathrm{b}}^2$ & 20 &  215,\,218 &13217 &11366 &44.1&49 &25 &217.597 &0.594&   $1.9\mathrm{E}4$ &100 \\%
  $I_{\mathrm{b}}^3$ & 30 &  110,\,113 &12993 &11346 &43.6 &48 &30& 110.193& 17.2&  $2.4\mathrm{E}7$& 100  \\%
  $I_{\mathrm{b}}^4$ & 40 &  137,\,140 &12767& 11324 &148.0&44 &25& 138.116 &0.16 &948 &100  \\%
  $I_{\mathrm{c}}$   & 40 &  150,\,153& 7900& 8629& 4.38  &50 &34& 151.133& 0.0197 &1 &1  \\%
  $I_{\mathrm{d}}$   & 40 &  -63,\,-61& 6507& 8106& 33.2  &45& 23& -61.733 &251.0 &  $4.9\mathrm{E}9$ &100  \\% \\
   \hline
   \end{tabular}\end{center}\label{table:stages_4_5_Fp}
\end{table}
Figure~\ref{fig:MILP_solutions}(a) illustrates  the chemical graph  $\Co^\dagger$  inferred 
 from   $I_{\mathrm{a}}$  with $(\underline{y}^*, \overline{y}^*) =(130,133)$  of  {\sc Fp}
  in Table~\ref{table:stages_4_5_Fp}.

 \begin{table}[h!]\caption{ Results of Stages~4 and 5 for {\sc OptR}.}  
 \begin{center}
 \begin{tabular}{@{}  c  c   c  r r r r r c  r r r     @{}}\hline                
 inst. & $n_\LB$ & $ \underline{y}^*,~\overline{y}^* $ & $\#$v~  &  $\#$c~~   &  
   {\small I-time}\!\! & $n$~  &  \!\!$\nint$  &   $\eta$   & 
                 {\small  D-time} &  {\small $\Co$-LB} &  {\small $\#\Co$}    \\ \hline 
   $I_{\mathrm{a}}$ & 30 &  81,\,83  & 10036 & 10396  & 4.32  & 47 & 26 & 81.587  & 0.066  & 2  & 2        \\%
  $I_{\mathrm{b}}^1$ & 10 &   -78,\,-76  &10570  & 8128 & 103.52 & 19  & 15 & -78.0 & 0.176 & 60 & 60      \\%
  $I_{\mathrm{b}}^2$ & 20 &   50,\,52  &12952  & 11231 & 41.2  & 49  &25 & 51.452  & 0.21 & 1616 & 100    \\%
  $I_{\mathrm{b}}^3$ & 30 &  30,\,32  &12722  &11201  & 37.3  &50  &25 & 31.261 & 0.186  &210  &100  \\%
  $I_{\mathrm{b}}^4$ & 40 &  116,\,118 & 12491 & 11171 & 17.7  & 49  &25  &117.918 & 0.883  &$3.7\mathrm{E}4$ & 100   \\%
  $I_{\mathrm{c}}$   & 40 &   -30,\,-28  &7885 & 8493  & 4.84  &49 & 33 & -29.163  &0.0159 & 1 & 1     \\%
  $I_{\mathrm{d}}$   & 40 &   40,\,42  &6495 &7976  & 10.5   &40 & 23 & 40.152 & 2.52 & $7.3\mathrm{E}5$ & 100  \\% \\
   \hline
   \end{tabular}\end{center}\label{table:stages_4_5_OptR}
\end{table}
 Figure~\ref{fig:MILP_solutions}(b) illustrates  the chemical graph  $\Co^\dagger$  inferred
 from   $I_{\mathrm{b}}^2$ with $(\underline{y}^*, \overline{y}^*) =(50, 52)$   of  {\sc OptR}
  in Table~\ref{table:stages_4_5_OptR}.

 \begin{table}[h!]\caption{ Results of Stages~4 and 5 for {\sc SfT}.}  
 \begin{center}
 \begin{tabular}{@{}  c  c  c  r r r r r c  r r r     @{}}\hline                
 inst. &  $n_\LB$ &  $ \underline{y}^*,~\overline{y}^* $ & $\#$v~  &  $\#$c~~   &  
   {\small I-time}\!\! & $n$~  &  \!\!$\nint$  & $\eta$   & 
                 {\small  D-time} &  {\small $\Co$-LB} &  {\small $\#\Co$}    \\ \hline
   $I_{\mathrm{a}}$ & 30 &   40,\,41 & 9288  &10207  &2.10  &41 & 22 & 40.231  &0.0638  &1  &1   \\%
  $I_{\mathrm{b}}^1$ & 10 &  28,\,29 & 7965 & 7599  &2.28 & 11 & 5  &28.488 & 0.00927  &1  &1   \\%
  $I_{\mathrm{b}}^2$ & 20 &  42,\,43  &9583  &10719 & 12.0  &48 & 25 & 42.998  &0.163  &200 & 100     \\%
  $I_{\mathrm{b}}^3$ & 30 & 36,\,37  &9326 & 10700 & 9.87  &42 & 25 & 36.358 & 9.39  &$9.2\mathrm{E}4$ & 100   \\%
  $I_{\mathrm{b}}^4$ & 40 & 43,\,44  &9069 & 10682  &13.4  &46 & 25 & 43.857  &0.123 & 116 & 100    \\%
  $I_{\mathrm{c}}$   & 40 &  44,\,45 & 7777 & 8378 & 4.37 & 45  &32 & 44.557  &0.0158 & 1  &1  \\%
  $I_{\mathrm{d}}$   & 40 &  39,\,40  &6385  &7857 & 6.72  &44 & 23  &39.55  &29.9 &  $2.8\mathrm{E}7$ &100  \\% \\
   \hline
   \end{tabular}\end{center}\label{table:stages_4_5_SfT}
\end{table}
Figure~\ref{fig:MILP_solutions}(c) illustrates  the chemical graph  $\Co^\dagger$  inferred 
 from  $I_{\mathrm{c}}$ with $(\underline{y}^*, \overline{y}^*) =(44, 45)$  of  {\sc  SfT}
  in Table~\ref{table:stages_4_5_SfT}.

In this experiment, we prepared several different types of instances:
 instances  $I_{\mathrm{a}}$ and $I_{\mathrm{c}}$ have restricted seed graphs,
the other  instances  have abstract seed graphs and
instances $I_{\mathrm{c}}$ and $I_{\mathrm{d}}$ have restricted set of fringe-trees.
From Tables~\ref{table:stages_4_5_Fp} and \ref{table:stages_4_5_SfT},
we observe that an instance with a large number of variables and constraints 
takes more running time than those with a smaller size in general.
All instances in this experiment are solved in a few seconds to around 150 seconds
with our MILP formulation.

\medskip \noindent
{\bf Stage~5.  } 
We executed Stage~5 to generate a more number of target chemical graphs $\Co^*$,
where we call a chemical graph $\Co^*$ a {\em chemical isomer} of
a target chemical graph $\Co^\dagger$ of a topological specification $\sigma$
if $f(\Co^*)=f(\Co^\dagger)$ and $\Co^*$ also satisfies the same topological specification $\sigma$.
We computed  chemical isomers  $\Co^*$ of 
each target chemical graph  $\Co^\dagger$ inferred in Stage~4.
We execute an  algorithm for generating chemical isomers of   $\Co^\dagger$
up to 100 when the number of all chemical isomers exceeds 100.
For this, we use a dynamic programming dynamic programming~\cite{ZAHZNA21}. 
The algorithm first decomposes $\Co^\dagger$ into a set of acyclic chemical graphs,
next replaces each acyclic chemical graph $T$ with another  acyclic chemical graph $T'$ that admits
the same feature vector as that of $T$
 and finally assembles the resulting acyclic chemical graphs
into a chemical isomer $\Co^*$ of $\Co^\dagger$. 
The algorithm can compute a lower bound 
on the total number of all chemical isomers $\Co^\dagger$
without generating all of them.

Tables~\ref{table:stages_4_5_Fp} and \ref{table:stages_4_5_SfT}  show
   the computational results of the experiment
in Stage~5 for the five properties, 
 where we denote the following:
\begin{enumerate}[nosep,  leftmargin=*]
\item[-]
 D-time: the running time (sec.) to execute the dynamic programming algorithm
 in Stage~5 to compute a lower bound on the number 
 of all chemical isomers  $\Co^*$ of  $\Co^\dagger$   
 and generate all (or up to 100) chemical isomers $\Co^*$;
 
\item[-]
 $\Co$-LB: a lower bound on the number of all chemical isomers $\Co^*$ of 
$\C^\dagger$, where $a\mathrm{E}b$ means $a\times 10^{b}$;~and

\item[-]
 $\#\Co$: the number of all (or up to 100) chemical isomers $\Co^*$ of  $\Co^\dagger$  
 generated in Stage~5.
\end{enumerate} 
  
From Tables~\ref{table:stages_4_5_Fp} and \ref{table:stages_4_5_SfT}, we observe  
 the running time and the number of generated isomers  in Stage~5. 
The chemical graph $\Co^\dagger$   in  $I_{\mathrm{b}^3}$
and $I_{\mathrm{d}}$ admits a large number of 
chemical isomers $\Co^*$ in some cases, 
where a lower bound $\Co$-LB  on the number of chemical isomers
is derived without generating all of them.  
For the other instances, the running time for generating up to 100 target chemical graphs
 in Stage~5 is less than one second.  
 For some chemical graph $\Co^\dagger$, no chemical isomer was found by our algorithm.
 This is because each acyclic chemical graph in the decomposition of $\Co^\dagger$
 has no alternative acyclic chemical graph than the original one. 
 For such an instance, we generate other desired chemical graphs
 by applying our new method of grid neighbor search. 

\medskip \noindent
{\bf Grid Neighbor Search.  } 
As a new building block of the framework of inferring chemical graphs,
we conducted an experiment of applying our grid neighbor search to generate other solutions
of an MILP in Stage~4.
We select the MILP such that a solution $\C^\dagger$ of the MILP 
admits at most two isomers $\C^*$ in Stage~5;
i.e.,    $I_{\mathrm{a}}$ with properties {\sc Fp}, {\sc OptR} and {\sc SfT}; 
$I_{\mathrm{b}}^1$ with properties {\sc Fp}  and {\sc SfT};  and 
$I_{\mathrm{c}}$ with properties {\sc Fp}, {\sc OptR}  and {\sc SfT}.
In this experiment, we set  
 $p_{\max}:=2$, 
 $\delta:=(0.1, 0.1)$, 
 $r:=(3,3)$ and 
 projection functions   $\theta_1$ and $\theta_2$ to be linear prediction functions
  for properties 
solubility ({\sc Sl}) and lipophilicity ({\sc Lp}), respectively,
constructed by Zhu et~al.~\cite{ZAHZNA21}.
The number of non-center grids in the neighbor $N(r)$ is $7\times 7-1=48$,
where the center grid $z=(0,0)$ is always feasible for the selected instances. 
For each instance, we check the feasibility of grids  $z \in N(r)$ 
in a non-decreasing order of $\max_{p\in [1,p_{\max}]}|z(p)|$.
For each feasible grid $z \in N(r)$, output a feasible solution $\C^\dagger_z$
of  the augmented MILP instance. 
We set a time limit for checking the feasibility of a grid to be 300 seconds,  
and we skip a grid when the corresponding MILP is not solved within the time limit.
We also discard any grid $z \in N(r)$
without testing the feasibility of $z$ if we find an infeasible grid $z'\in N(r)$
with $z'\prec z$.

Tables~\ref{table:grid}    shows
   the computational results of the experiment
in the grid neighbor search for the eight instances, 
 where we denote the following:
\begin{enumerate}[nosep,  leftmargin=*]
\item[-]
instance: a pair $(I,\pi)$ of topological specification $I$ and
a property $\pi$ in the tested instance;

\item[-] $n$: the number of non-hydrogen atoms in the tested instance;  

\item[-] \#feasible:  the number of  non-center grids in $N(r)$
that are found to be feasible during the search procedure; 

\item[-] \#solutions:   the number of chemical graphs
obtained from the feasible non-center grids in $N(r)$,
where different feasible girds provide the same solutions
and \#feasible $>$ \#solutions holds in such a case; 

\item[-] \#infeasible: the number of  non-center grids in $N(r)$
that are found to be infeasible during the search procedure; 

\item[-] \#ignored:  the number of  non-center grids $z\in N(r)$
that are ignored due to an infeasible grid $z'z\in N(r)$ with $z'\prec z$
during the search procedure; 

\item[-] \#time  out: the number of  non-center grids in $N(r)$
such that the time for feasibility check exceeds
the time limit of 300 seconds during the search procedure; 

  \item[-] time: the average time for checking the feasibility of
 a non-center grid $z\in N(r)$ whose feasibility can be detected within the time limit.
\end{enumerate} 

 \begin{table}[h!]\caption{ Results of Grid Neighbor Search.}  
 \begin{center}
 \begin{tabular}{@{}  c  c c c    c c c c     @{}}\hline            
 instance & $n$ &  \#feasible & \#solutions &  \#infeasible & \#ignored & \#time out & time \\ \hline
  $(I_{\mathrm{a}}$,{\sc Fp})   & 42 & 33 & 33  & 0 & 0 & 15 &   72.96  \\
  $(I_{\mathrm{a}}$,{\sc OptR}) & 47  & 30 & 30 & 1  & 2 & 15 &  44.00  \\
  $(I_{\mathrm{a}}$,{\sc SfT})     & 41 & 45  &  45 & 0 & 0 & ~3 &  11.91  \\
  $(I_{\mathrm{b}}^1$,{\sc Fp})     & 10  & 21 &21  & 3  & 3 & 21  & 69.48  \\
  $(I_{\mathrm{b}}^1$,{\sc SfT})    & 11  &  22 & 22 & 0 & 0 & 26 &  30.78  \\
  $(I_{\mathrm{c}}$,{\sc Fp})      &  50 & 33  & 33 & 0 & 0 & 15  &  71.84  \\
  $(I_{\mathrm{c}}$,{\sc OptR})     & 49 & 25 & 25 & 1 & 1 & 21 &   13.29  \\
  $(I_{\mathrm{c}}$,{\sc SfT})      & 45 & 17 &17  & 3  & 4  & 24  &  65.85  \\\hline
   \end{tabular}\end{center}\label{table:grid}  
\end{table}

From Tables~\ref{table:grid}, we observe that our new method of grid neighbor search
successfully infers other solutions than the chemical graphs $\C^\dagger$
 inferred by the standard Stage~4
even though Stage~5 could not find many chemical isomers $\C^*$ of $\C^\dagger$.
The branch-and-bound method for solving an MILP sometimes takes
an extremely large execution time for the same size of instances. 
We introduce a time limit to bound an entire running time to skip such instances
during an execution of testing the feasibility of grids in the neighbor $N(r)$.
From Tables~\ref{table:grid}, we see that at least around a half number of grids in $N(r)$
were feasible and provided new solutions.

%\clearpage  

\section{Concluding Remarks}\label{sec:conclude}%%%%%%%%%%%%%%%%%%
%\section{Discussions \& Conclusions}\label{sec:conclude}%%%%%%%%%%%%%%%%%%
 
In this paper, we designed a procedure for generating chemical graphs
as a new building block 
in Stage~4 of the framework for inferring a desired chemical graph.
The main task of Stage~4 is to find a feasible solution $\C^\dagger$ of an MILP that
represents a feature function and a topological specification.
In the framework, isomers $\C^*$ of $\C^\dagger$ are generated 
in Stage~5 by a dynamic programming algorithm.
However, the number of isomers of $\C^\dagger$ is sometimes small.
Our new procedure searches the neighbor of $\C^\dagger$ in a search space
 defined with a set of linear functions.
We divide the neighbor of $\C^\dagger$ into subspaces and
solve the MILP in Stage~4 for each subspace 
imposing a set of linear constraints that represents the subspace.
From the results of our computational experiments, we observe that
an additional number of  solutions $\C^*$
 can be found in the neighbor of $\C^\dagger$ 
by our new procedure. 

% It is left as a future work to use other learning methods such as
% random forest, graph convolution and an ensemble method 
% in Stages~3 and 4 in the framework. 

\clearpage
 \appendix
\centerline{\bf\LARGE Appendix}

\section{A Full Description of Descriptors}\label{sec:descriptor}%%%%%%%%%

Associated with the two functions 
$\alpha$ and $\beta$ in a chemical graph $\Co=(H,\alpha,\beta)$,
we introduce   functions  
 $\ac: V(E)\to (\Lambda\setminus\{\ttH\})\times (\Lambda\setminus\{\ttH\})\times [1,3]$, 
 $\cs: V(E)\to (\Lambda\setminus\{\ttH\})\times [1,6]$ and
$\ec: V(E)\to ((\Lambda\setminus\{\ttH\})\times [1,6])\times ((\Lambda\setminus\{\ttH\})\times [1,6])\times [1,3]$
in the following. 

 To represent  a feature of the exterior  of  $\Co$, 
  a  chemical rooted tree in $\mathcal{T}(\Co)$ is
  called a {\em fringe-configuration} of $\Co$. 

We also represent leaf-edges in the exterior of $\Co$.
For a leaf-edge $uv\in E(\anC)$ with $\deg_{\anC}(u)=1$, we define
the {\em adjacency-configuration} of $e$ to be an ordered tuple
$(\alpha(u),\alpha(v),\beta(uv))$. 
Define 
\[ \Gac^\lf\triangleq \{(\ta,\tb,m)\mid \ta,\tb\in\Lambda, 
m\in[1,\min\{\val(\ta),\val(\tb)\}]\} \]
as a set of possible adjacency-configurations for leaf-edges. 

To  represent a feature of an interior-vertex $v\in V^\inte(\Co)$ such that
$\alpha(v)=\ta$  and  $\deg_{\anC}(v)=d$
(i.e., the number of non-hydrogen atoms adjacent to $v$ is $d$) 
   in a chemical   graph  $\Co=(H,\alpha,\beta)$,
 we use  a pair $(\ta, d)\in (\Lambda\setminus\{{\tt H}\})\times [1,4]$,
 which we call the {\em chemical symbol} $\cs(v)$ of the vertex $v$.
 We treat $(\ta, d)$ as a single symbol $\ta d$,  and  
define $\Ldg$   to be  the set of all chemical symbols
$\mu=\ta d\in  (\Lambda\setminus\{{\tt H}\})\times [1,4]$.  
% For a notational convenience,
% we write a chemical symbol $(\ta, i)$ (resp., $(\ta, i+j)$) 
% as $\ta i$ (resp., $\ta\{i+j\}$). 

We define a method for featuring interior-edges  as follows.
Let $e=uv\in E^\inte(\Co)$  be 
 an interior-edge $e=uv\in E^\inte(\Co)$ 
 such that $\alpha(u)=\ta$, $\alpha(v)=\tb$ and $\beta(e)=m$ 
   in a chemical graph  $\Co=(H,\alpha,\beta)$.
To feature this edge $e$, 
 we use a tuple $(\ta,\tb,m)\in (\Lambda\setminus\{{\tt H}\})
    \times (\Lambda\setminus\{{\tt H}\})\times [1,3]$,
 which we call the {\em adjacency-configuration} $\ac(e)$ of the edge $e$. 
 We introduce a total order $<$ over the elements in $\Lambda$
 to distinguish  between $(\ta,\tb, m)$ and $(\tb,\ta, m)$ 
 $(\ta\neq \tb)$ notationally.
 For a tuple  $\nu=(\ta,\tb, m)$,
 let $\overline{\nu}$ denote the tuple $(\tb,\ta, m)$.

Let $e=uv\in E^\inte(\Co)$  be 
an  interior-edge $e=uv\in E^\inte(\Co)$ 
 such that $\cs(u)=\mu$, $\cs(v)=\mu'$ and $\beta(e)=m$ 
   in a chemical  graph  $\Co=(H,\alpha,\beta)$.
To feature this edge $e$, 
 we use a tuple $(\mu,\mu',m)\in \Ldg\times \Ldg\times [1,3]$, 
 which we call  the {\em edge-configuration} $\ec(e)$ of the edge $e$. 
 We introduce a total order $<$ over the elements in $\Ldg$
 to distinguish between $(\mu,\mu', m)$ and $(\mu', \mu, m)$ 
 $(\mu \neq \mu')$ notationally. 
 For a tuple  $\gamma=(\mu,\mu', m)$,
 let $\overline{\gamma}$ denote the tuple $(\mu', \mu, m)$. 

Let $\pi$ be a chemical property for which we will construct
a prediction function $\eta$ from a feature
vector $f(\Co)$ of a chemical graph $\Co$ 
to a predicted value $y\in \mathbb{R}$
for the  chemical property of $\Co$.

We first choose a set $\Lambda$ of chemical elements
 and then collect a data set  $D_{\pi}$ of
  chemical compounds  $C$ whose 
  chemical elements belong to $\Lambda$,
  where we regard  $D_{\pi}$ as a set of chemical graphs $\Co$
  that represent the chemical compounds $C$  in  $D_{\pi}$.
To define the interior/exterior of 
chemical graphs  $\Co\in D_{\pi}$,
we  next choose a branch-parameter ${\rho}$, where
 we recommend ${\rho}=2$.  
 
Let $\Lambda^\inte(D_\pi)\subseteq \Lambda$ 
(resp., $\Lambda^\ex(D_\pi)\subseteq \Lambda$)
denote the set  of chemical elements  used in
the set $V^\inte(\Co)$ of interior-vertices
(resp., the set $V^\ex(\Co)$ of  exterior-vertices) of $\Co$
 over all chemical graphs $\Co\in D_\pi$, 
and $\Gamma^\inte(D_\pi)$
denote the set of edge-configurations used in
the set $E^\inte(\Co)$  of interior-edges in $\Co$
 over all chemical graphs $\Co\in D_\pi$. 
Let $\mathcal{F}(D_\pi)$ denote the set of
chemical rooted trees $\psi$  
r-isomorphic to a chemical rooted tree in $\mathcal{T}(\Co)$
  over all chemical graphs $\Co\in D_\pi$,
  where possibly a chemical rooted tree $\psi\in \mathcal{F}(D_\pi)$
  consists of a single chemical element $\ta\in \Lambda\setminus \{{\tt H}\}$.
  
We define an integer encoding of a finite set $A$ of elements
to be a bijection $\sigma: A \to [1, |A|]$, 
where we denote by $[A]$   the set $[1, |A|]$ of integers.
Introduce  an integer coding of each of the   sets 
$\Lambda^\inte(D_\pi)$, $\Lambda^\ex(D_\pi)$, 
$\Gamma^\inte(D_\pi)$ and $\mathcal{F}(D_\pi)$. 
Let $[\ta]^\inte$  
(resp., $[\ta]^\ex$)  denote   
the coded integer of  an element $\ta\in \Lambda^\inte(D_\pi)$
(resp., $\ta\in \Lambda^\ex(D_\pi)$),  
$[\gamma]$   denote  
the coded integer of  an element $\gamma$ in $\Gamma^\inte(D_\pi)$
and 
$[\psi]$   denote  an element $\psi$ in $\mathcal{F}(D_\pi)$. 
 
 Over 99\% of  chemical compounds $\Co$ with up to
  100 non-hydrogen atoms in  PubChem  have degree at most 4
 in the hydrogen-suppressed graph $\anC$~\cite{AZSSSZNA20}. 
We assume that a chemical graph $\Co$
 treated in this paper satisfies  $\deg_{\anC}(v)\leq 4$
in the hydrogen-suppressed graph $\anC$.
 
In our model, we  use an integer 
  $\mathrm{mass}^*(\ta)=\lfloor 10\cdot \mathrm{mass}(\ta)\rfloor$, 
 for each $\ta\in \Lambda$.
% and assume that
% each chemical element $\ta\in \Lambda$ has a unique 
% valence  $\val(\ta)\in [1,4]$.  
 
  We define the {\em feature vector} $f(\C)$ 
  of a  chemical graph $\C=(H,\alpha,\beta)\in D_{\pi}$ 
  to be a vector that consists of the following  
non-negative integer descriptors $\dcp_i(\C)$, $i\in [1,K]$, where 
% 4+4\times 2 + 2\times 3 + 1 +1 =4+8+6 +1 +1 = 20
$K = 14+ |\Lambda^\inte(D_\pi)|+|\Lambda^\ex(D_\pi)|
         +|\Gamma^\inte(D_\pi)|+|\mathcal{F}(D_\pi)|+|\Gac^\lf|$. 

%  $\Lambda^+=\{ \ttN \}$ and  $\Lambda^-=\{ \ttO \}$. 
% $N^-, O^-, P^-, B^-, Al^-$     $N^+$

\begin{enumerate}  % [nosep] %[nosep,  leftmargin=*]% 
\item   
$\dcp_1(\C)$: the number  $|V(H)|-|\VH|$ of non-hydrogen atoms  in  $\C$.  
 
\item   
$\dcp_2(\C)$: the rank $\mathrm{r}(\C)$ of   $\C$.  

\item 
$\dcp_3(\C)$:  the number $|V^\inte(\C)|$ of interior-vertices in  $\C$.
  
\item 
$\dcp_4(\C)$: 
the average $\overline{\mathrm{ms}}(\C)$ of mass$^*$ 
over all atoms in $\C$; \\
 i.e., $\overline{\mathrm{ms}}(\C)\triangleq 
 \frac{1}{|V(H)|}\sum_{v\in V(H)}\mathrm{mass}^*(\alpha(v))$. 

\item 
$\dcp_i(\C)$,  $i=4+d,   d\in [1,4]$: 
the number $\dg_d^{\oH} (\C)$ 
 of non-hydrogen vertices $v\in V(H)\setminus \VH$
 of degree $\deg_{\anC}(v)=d$
 in the hydrogen-suppressed chemical graph $\anC$.  
 
\item 
$\dcp_i(\C)$,  $i=8+d,   d\in [1,4]$: 
the number $\dg_d^\inte(\C)$
 of interior-vertices of interior-degree  $\deg_{\C^\inte}(v)=d$
  in the interior $\C^\inte=(V^\inte(\C),E^\inte(\C))$ of  $\C$. 
  
%\item $\dcp_i(\C)$,  $i=11+d,   d\in [0,3]$: 
%the number $\hydg_d(\Co)$ of vertices in $G$ of hydro-degree  $\hyddeg (v)=d$. 
   
\item $\dcp_i(\C)$, $i=12+m$,  $m\in[2,3]$: 
the number $\bd_m^\inte(\C)$
 of  interior-edges with bond multiplicity $m$ in  $\C$; 
 i.e., $\bd_m^\inte(\C)\triangleq \{e\in E^\inte(\C)\mid \beta(e)=m\}$.

\item $\dcp_i(\C)$, $i=14+[\ta]^\inte$, 
 $\ta\in \Lambda^\inte(D_\pi)$: 
 the frequency $\na_\ta^\inte(\C)=|V_\ta(\C)\cap V^\inte(\C) |$ 
 of chemical element $\ta$ in
 the set $V^\inte(\C)$ of  interior-vertices in  $\C$. 
 
\item $\dcp_i(\C)$, 
$i=14+|\Lambda^\inte(D_\pi)|+[\ta]^\ex$, 
 $\ta\in \Lambda^\ex(D_\pi)$: 
 the frequency $\na_\ta^\ex(\C)=|V_\ta(\C)\cap V^\ex(\C) |$
  of chemical element $\ta$ in
 the set $V^\ex(\C)$ of  exterior-vertices in  $\C$. 
 
\item $\dcp_i(\C)$, 
$i=14+|\Lambda^\inte(D_\pi)|+|\Lambda^\ex(D_\pi)|+ [\gamma]$, 
$\gamma \in \Gamma^\inte(D_\pi)$: 
the frequency $\ec_{\gamma} (\Co)$ of edge-configuration $\gamma$
in the set $E^\inte(\C)$ of interior-edges   in  $\C$.

\item $\dcp_i(\C)$, 
$i= 14+|\Lambda^\inte(D_\pi)|+|\Lambda^\ex(D_\pi)|
+ |\Gamma^\inte(D_\pi)|+ [\psi]$,  
 $\psi \in \mathcal{F}(D_\pi)$: 
the frequency $\fc_{\psi}(\C)$ of fringe-configuration $\psi $
in the set of ${\rho}$-fringe-trees in  $\C$. 

\item $\dcp_i(\C)$, 
$i= 14+|\Lambda^\inte(D_\pi)|+|\Lambda^\ex(D_\pi)|
+ |\Gamma^\inte(D_\pi)|+|\mathcal{F}(D_\pi)|+ [\nu]$,  
 $\nu \in \Gac^\lf$: 
the frequency $\ac_{\nu}^\lf(\C)$ of adjacency-configuration $\nu$
in the set of leaf-edges in  $\anC$. %\newone
\end{enumerate}

\section{Specifying Target Chemical Graphs}\label{sec:specification} 

Given a prediction function $\eta$ and 
a target value $y^*\in \mathbb{R}$, 
we call a chemical graph $\C^*$ such that $\eta(x^*)=y^*$
for the feature vector $x^*=f(\C^*)$ a {\em target chemical graph}.
This section  presents a set of rules for 
 specifying  topological substructure
  of a target chemical graph in a flexible way in Stage~4.

We first describe how to reduce a chemical graph $\C=(H,\alpha,\beta)$ into
an abstract form based on which our specification rules will be defined.
To illustrate the reduction process,
we use the chemical graph $\C=(H,\alpha,\beta)$
such that $\anC$ is given in Figure~\ref{fig:example_chemical_graph}.
 
\begin{figure}[h!] \begin{center}
\includegraphics[width=.65\columnwidth]{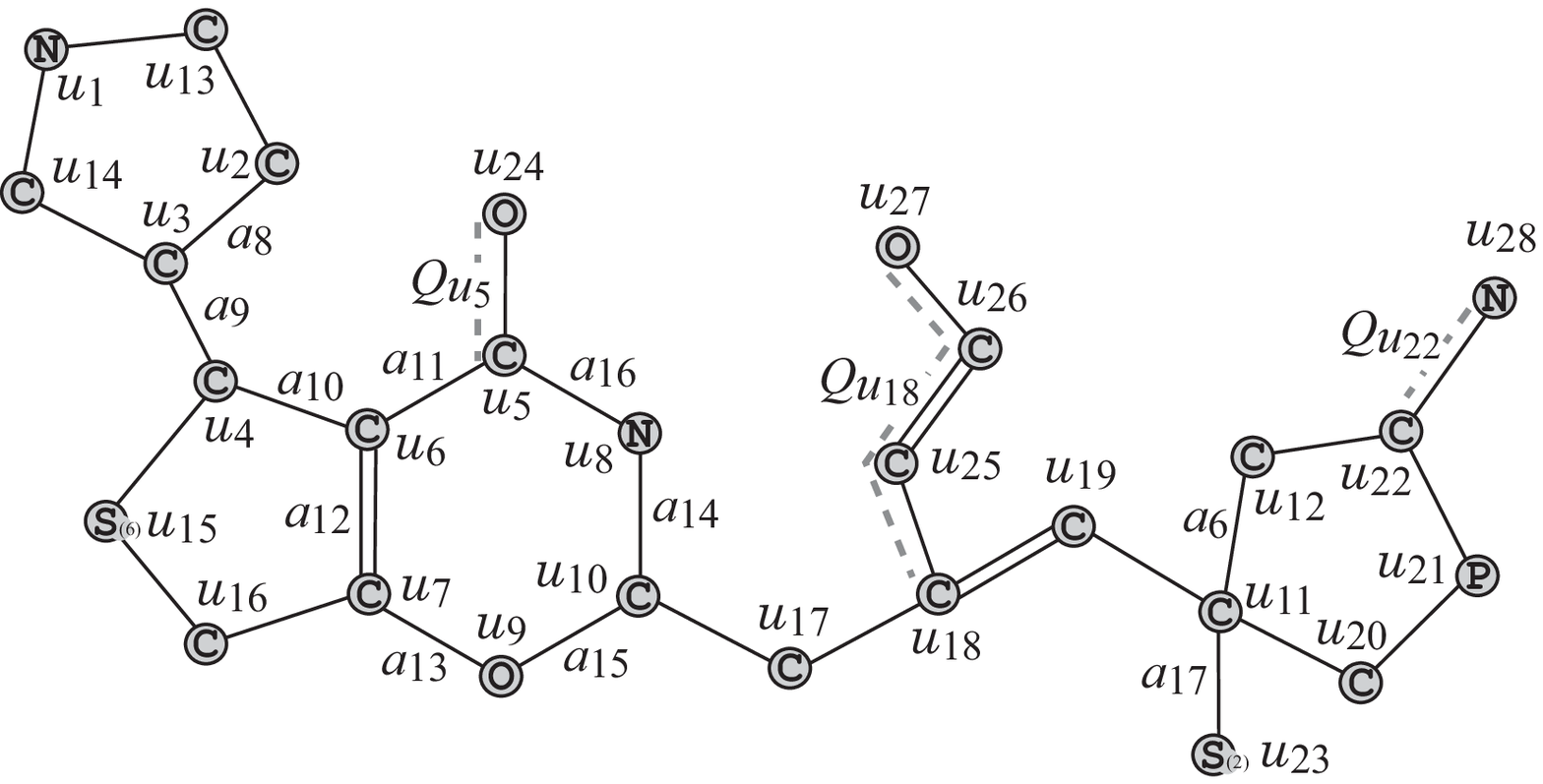}
\end{center} \caption{The interior $H^\inte$ of
chemical graph $\C$ with $\anC$ 
  in Figure~\ref{fig:example_chemical_graph} for ${\rho}=2$.
}
\label{fig:specification_example_interior} \end{figure}

\begin{figure}[h!] \begin{center}
\includegraphics[width=.98\columnwidth]{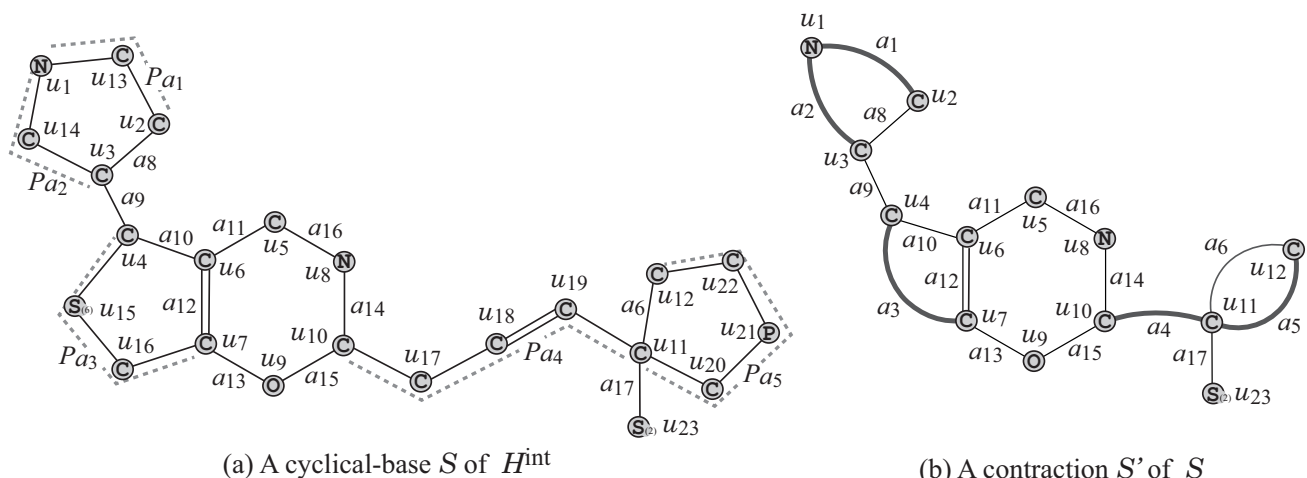}
\end{center} \caption{
(a) A cyclical-base  
$S=H^\inte- \bigcup_{u\in \{u_5,u_{18},u_{22}\}}(V(Q_u)\setminus \{u\})$
of the interior  $H^\inte$ in Figure~\ref{fig:specification_example_interior};
(b) A contraction $S'$ of  $S$ for a pure path set 
 $\mathcal{P}=\{P_{a_1},P_{a_2},\ldots,P_{a_5}\}$ 
%P_1=(u_1,u_{23},u_{2}), 
% P_2=(u_{1},u_{14},u_{3}),
% P_3=(u_{4},u_{15},u_{16},u_{7}), 
% P_4=(u_{10},u_{17},u_{18},u_{19},u_{11}),
% P_5=(u_{11},u_{20},u_{21},u_{22},u_{12})\}$  
in (a),
where a new edge obtained by contracting a pure path is depicted
with a thick line.   
}
\label{fig:specification_example_R2_3} \end{figure}

 \begin{enumerate}%[leftmargin=*]
 \item[R1] {\bf Removal of all ${\rho}$-fringe-trees: } 
The interior $H^\inte=(V^\inte(\C),E^\inte(\C))$ of $\C$ 
is obtained by removing the non-root vertices of 
each ${\rho}$-fringe-trees $\C[u]\in\mathcal{T}(\C), u\in V^\inte(\C)$. 
Figure~\ref{fig:specification_example_interior} illustrates
the interior $H^\inte$ of 
chemical graph $\C$ with ${\rho}=2$
  in Figure~\ref{fig:example_chemical_graph}. 
  
 \item[R2] {\bf Removal of some leaf paths: } 
 We call a $u,v$-path $Q$ in $H^\inte$  a {\em leaf path} if 
  vertex $v$ is a leaf-vertex of $H^\inte$
  and the degree of each internal vertex of $Q$  in $H^\inte$  is 2,
  where we regard that $Q$ is rooted at vertex $u$. 
A connected subgraph $S$ of the interior $H^\inte$ of $\C$  
is called a {\em cyclical-base}
if $S$ is obtained from $H$
by removing the vertices in $V(Q_u)\setminus \{u \}, u\in X$ 
for a subset $X$ of interior-vertices  and a set  $\{Q_u \mid u\in X\}$ of leaf 
 $u,v$-paths $Q_u$  such that    
 no two paths $Q_u$ and $Q_{u'}$ share a vertex.
Figure~\ref{fig:specification_example_R2_3}(a) illustrates
a cyclical-base  $S=H^\inte- \bigcup_{u\in X}(V(Q_u)\setminus \{u\})$
of the interior  $H^\inte$  
for a set 
$\{Q_{u_5}=(u_5,u_{24}), 
     Q_{u_{18}}=(u_{18},u_{25},u_{26},u_{27}),
     Q_{u_{22}}=(u_{22},u_{28})\}$ of leaf  paths 
in Figure~\ref{fig:specification_example_interior}.  

 \item[R3] {\bf Contraction of some pure paths: } 
 A path in $S$ is called {\em pure} 
 if  each internal vertex of the path  is of degree 2. 
 Choose a set $\mathcal{P}$ of several pure paths in $S$ 
 so that no two paths share  vertices except for their end-vertices. 
 A graph $S'$ is called a {\em contraction} of a graph $S$
  (with respect to $\mathcal{P}$) 
 if $S'$ is obtained from $S$ by replacing 
 each pure $u,v$-path  with a single edge $a=uv$,
 where $S'$ may contain multiple edges between the same pair of adjacent vertices.
Figure~\ref{fig:specification_example_R2_3}(b) illustrates
a contraction $S'$ obtained from 
the chemical graph  $S$
by contracting each $uv$-path $P_a\in  \mathcal{P}$ into a new edge $a=uv$,
where $a_1=u_1 u_{2},  a_2=u_1 u_{3},  a_3=u_4 u_{7}, a_4=u_{10}u_{11}$
and $a_5=u_{11}u_{12}$ and 
 $\mathcal{P}=\{
 P_{a_1}=(u_1,u_{13},u_{2}), 
 P_{a_2}=(u_{1},u_{14},u_{3}),
 P_{a_3}=(u_{4},u_{15},u_{16},u_{7}), 
 P_{a_4}=(u_{10},u_{17},u_{18},u_{19},u_{11}),
 P_{a_5}=(u_{11},u_{20},u_{21},u_{22},u_{12})\}$ of pure paths 
in Figure~\ref{fig:specification_example_R2_3}(a). 
\end{enumerate}

We will define a set of rules so that 
a chemical graph can be obtained 
from a graph (called a seed graph in the next section)
by applying processes R3 to R1 in a reverse way. 
We specify topological substructures of a target chemical graph
with a tuple  $(\GC,\sint,\sce)$  called  a {\em target specification}
defined under the set of the following rules. 

\subsection*{Seed Graph}%%%%%%%%%%%%%%%%%%%%

A  {\em seed graph} $\GC=(\VC,\EC)$ is defined
to be a graph (possibly with multiple edges) such that 
the edge set $\EC$ consists of four sets 
$\Et$, $\Ew$, $\Ez$ and $\Eew$, 
where each of them can be empty.
A seed graph plays a role of the most abstract form $S'$ in R3.  
Figure~\ref{fig:specification_example_1}(a) illustrates an example of a seed graph
$\GC$ with $\mathrm{r}(\GC)=5$,   
where $\VC=\{u_1,u_2,\ldots,u_{12},u_{23}\}$, 
$\Et=\{a_1,a_2,\ldots,a_5\}$, 
$\Ew=\{a_6\}$,
$\Ez=\{a_7\}$ and 
$\Eew=\{a_8,a_9,\ldots,a_{16}\}$.

 A {\em subdivision} $S$ of $\GC$  
is a graph constructed from a seed graph $\GC$ 
according to the following rules:
\begin{enumerate}[leftmargin=*]
\item[-]
Each edge $e=uv\in \Et$ is replaced
with a $u,v$-path $P_e$ of length at least 2;

\item[-] 
Each edge $e=uv\in \Ew$ is replaced
with a $u,v$-path $P_e$ of length at least 1
(equivalently $e$ is directly used or replaced with
a $u,v$-path $P_e$ of length at least 2);

\item[-] 
Each edge $e\in \Ez$ is either used or discarded, where 
 $\Ez$ is required to be chosen as a non-separating edge subset of
 $E(\GC)$ since otherwise the connectivity of a final chemical graph $\Co$
 is not guaranteed; 
$\mathrm{r}(\Co)= \mathrm{r}(\GC)-|E'|$ holds
for a subset $E'\subseteq \Ez$ of edges discarded 
in a  final chemical graph $\Co$; 
and 

\item[-]
Each edge $e\in \Eew$ is always used directly. 
\end{enumerate}

We allow a possible elimination of edges in $\Ez$ as an optional rule
in constructing a target chemical graph from a seed graph, 
even though such an operation has 
not been included in the process R3. 
A subdivision  $S$ plays a role of a cyclical-base   in R2. 
A target chemical graph $\C=(H,\alpha,\beta)$ will contain  $S$  as a subgraph
of the interior $H^\inte$ of $\C$.

% \clearpage 

\subsection*{Interior-specification}%%%%%%

A graph $H^*$ that serves as the interior $H^\inte$ of
a target chemical graph $\C$ will be constructed as follows.
First construct a subdivision  $S$ of a seed graph $\GC$ 
by replacing each edge $e=u u'\in \Et\cup\Ew$
with a pure $u,u'$-path $P_e$.
Next construct a supergraph $H^*$ of $S$ by 
attaching a leaf path $Q_v$ at each vertex $v\in \VC$ or
at an internal vertex $v\in V(P_e)\setminus\{u,u'\}$ 
of each pure $u,u'$-path $P_e$ for some edge $e=uu'\in \Et\cup\Ew$,
where possibly $Q_v=(v), E(Q_v)=\emptyset$ 
(i.e., we do not attach any new edges to $v$).
We introduce the following rules for specifying
 the size of $H^*$, the length $|E(P_e)|$  of
a pure path  $P_e$,  the length $|E(Q_v)|$ of
a   leaf path $Q_v$, the number of  leaf paths $Q_v$
and a bond-multiplicity of each interior-edge,
where we call the set of prescribed constants  
 an  {\em interior-specification}   $\sint$: 
\begin{enumerate}[leftmargin=*]
 \item[-]
  Lower and upper bounds $\nint_\LB, \nint_\UB\in \mathbb{Z}_+$ 
  on   the number of interior-vertices 
of a target chemical graph~$\C$. 
  
\item[-] 
For each edge $e=u u'\in \Et\cup\Ew$, 
\begin{description} 
\item[]
 a lower bound $\ell_{\LB}(e)$ and 
 an upper bound $\ell_{\UB}(e)$  on the length $|E(P_e)|$ of
 a pure $u,u'$-path $P_e$. 
(For a notational convenience, set 
$\ell_\LB(e):=0$, $\ell_\UB(e):=1$, $e\in \Ez$ and
$\ell_\LB(e):=1$, $\ell_\UB(e):=1$, $e\in \Eew$.)
   
\item[]  
 a lower bound $\bl_{\LB}(e)$ and 
 an upper bound $\bl_{\UB}(e)$ on the number of leaf paths $Q_v$ attached 
 at  internal vertices $v$ of a pure $u,u'$-path $P_e$.   

\item[] 
 a lower bound $\ch_{\LB}(e)$ and 
 an upper bound $\ch_{\UB}(e)$  on the maximum 
 length  $|E(Q_v)|$ of a leaf path $Q_v$ attached  
 at an internal vertex $v\in V(P_e)\setminus\{u,u'\}$ 
 of a pure $u,u'$-path $P_e$.   
\end{description} 

\item[-]
For each vertex $v\in \VC$, 
\begin{description} 
\item[]
 a lower bound $\ch_{\LB}(v)$ and 
 an upper bound $\ch_{\UB}(v)$  on  
 the number of leaf paths $Q_v$ attached to $v$,
 where $0\leq \ch_{\LB}(v)\leq \ch_{\UB}(v)\leq 1$.
 
\item[]
 a lower bound $\ch_{\LB}(v)$ and 
 an upper bound $\ch_{\UB}(v)$  on the
 length $|E(Q_v)|$ of a leaf path $Q_v$ attached to $v$. 
\end{description}  

\item[-] 
For each edge $e=u u'\in \EC$, 
a lower bound $\bd_{m, \LB}(e)$ 
and an  upper bound $\bd_{m, \UB}(e)$  on
the number of edges with bond-multiplicity $m\in [2,3]$ in
$u,u'$-path $P_e$, where we regard $P_e$, $e  \in \Ez\cup \Eew$ 
as single edge $e$.
\end{enumerate}

We call a graph $H^*$ that satisfies an interior-specification $\sint$
a {\em $\sint$-extension of $\GC$}, 
where the bond-multiplicity of each edge has been determined.

Table~\ref{table:interior-spec}  shows an example of 
an interior-specification  $\sint$ to the seed graph  $\GC$ in 
Figure~\ref{fig:specification_example_1}. 

\begin{table}[h!]\caption{Example~1 of an interior-specification  $\sint$. }
% \begin{center}
\begin{tabular}{ |  c | c |  } \hline 
$\nint_\LB=20$ & $\nint_\UB = 28$ \\\hline 
\end{tabular}

 \begin{tabular}{ |  c | c c c c c c |  } \hline
                        & $a_1$ &  $a_2$ &   $a_3$ &   $a_4$ &   $a_5$ &   $a_6$   \\\hline
 $\ell_\LB(a_i)$&  2 &  2 &  2 & 3 &  2 &  1 \\ \hline
 $\ell_\UB(a_i)$&  3 & 4 &  3 & 5 & 4 &  4 \\\hline
 $\bl_\LB(a_i)$&  0 &  0 &   0 & 1 &  1 &   0 \\ \hline
 $\bl_\UB(a_i)$&  1 & 1 &   0 & 2 & 1 &   0 \\\hline
 $\ch_\LB(a_i)$&  0 &  1 & 0 & 4 &  3 &  0 \\ \hline
 $\ch_\UB(a_i)$&  3 & 3 &  1 & 6 & 5 &  2 \\\hline
\end{tabular} 

\begin{tabular}{ |  c | c c c c c c   c c c c  c c c |  } \hline
                        & $u_1$ &  $u_2$ &   $u_3$ &   $u_4$ &   $u_5$ &   $u_6$ 
                       & $u_7$ &   $u_8$ &   $u_9$ &   $u_{10}$ &   $u_{11}$ 
                       &   $u_{12}$ &   $u_{23}$ \\\hline 
 $\bl_\LB(u_i)$&  0 &  0 &   0 & 0 &  0 &   0
                       & 0 &   0 &  0 &   0 &  0 &  0 &  0 \\ \hline
 $\bl_\UB(u_i)$&  1 & 1 &   1 & 1 & 1 &   0
                       & 0 &   0 &  0 &   0 &  0 &  0 &  0\\\hline
 $\ch_\LB(u_i)$&  0 &  0 &   0 & 0 &  1 &   0
                       & 0 &   0 &  0 &   0 &  0 &  0 &  0 \\ \hline
 $\ch_\UB(u_i)$&  1 & 0 &   0 & 0 & 3 &   0
                       & 1 &   1 &  0 &   1 &  2 & 4 &  1 \\\hline
\end{tabular} 

\begin{tabular}{ |  c | c c c c c c   c c c c c c  c c c c c |  } \hline
                               & $a_1$ &  $a_2$ &   $a_3$ &   $a_4$ &   $a_5$ &   $a_6$ 
                               & $a_7$ &  $a_8$ &   $a_9$ &   $a_{10}$ &   $a_{11}$ &   $a_{12}$ 
                               & $a_{13}$ &   $a_{14}$ &   $a_{15}$ &   $a_{16}$ &   $a_{17}$  \\\hline
 $\bd_{2, \LB}(a_i)$ &  0    &  0 &   0 & 1 &  0 &   0
                                &  0   &  0 &  0 & 0 &  0 &   1
                                &  0    &  0 &   0 & 0     & 0  \\ \hline
 $ \bd_{2, \UB}(a_i)$&  1    & 1 &   0 & 2  & 2 &   0  
                                &  0    & 0&   0 & 0 &  0 &   1
                                &  0    &  0 &   0 & 0   & 0   \\ \hline
 $\bd_{3, \LB}(a_i)$ &  0    &  0 &   0 & 0 &  0 &   0
                                &  0   &  0 &  0 & 0 &  0 &   0
                                &  0    &  0 &   0 & 0   & 0   \\ \hline
 $ \bd_{3, \UB}(a_i)$&  0    & 0 &   0 & 0  & 1 &   0 
                                &  0    &  0 &   0 & 0 &  0 &   0
                                &  0    &  0 &   0 &  0    & 0   \\ \hline
\end{tabular} %\end{center}
\label{table:interior-spec}  
\end{table}

\begin{figure}[h!] \begin{center}
\includegraphics[width=.58\columnwidth]{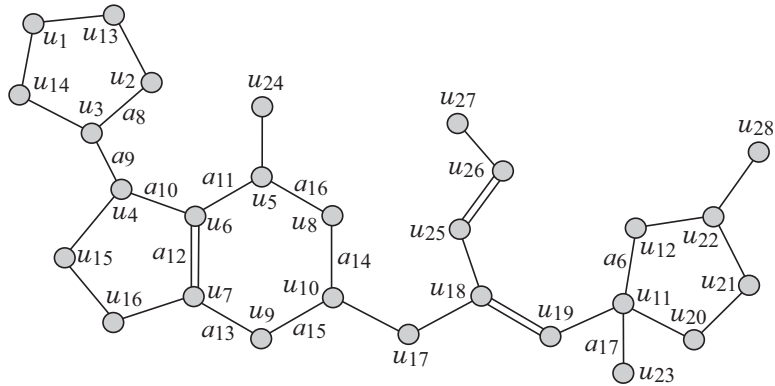}
\end{center} \caption{
An illustration of a graph 
$H^*$ that is obtained from  the seed graph  $\GC$ in 
Figure~\ref{fig:specification_example_1}
under the interior-specification  $\sint$ in 
Table~\ref{table:interior-spec},
where the vertices newly introduced by pure paths $P_{a_i}$
and leaf paths $Q_{v_i}$ are depicted with white squares and circles,
respectively.    }
\label{fig:specification_example_3} \end{figure}

Figure~\ref{fig:specification_example_3} illustrates an example of 
an $\sint$-extension $H^*$ of seed graph  $\GC$ in 
Figure~\ref{fig:specification_example_1}
under the interior-specification  $\sint$ in 
Table~\ref{table:interior-spec}.  

%\clearpage 

\subsection*{Chemical-specification}%%%%%%
 
 Let $H^*$ be a graph that serves as 
 the interior $H^\inte$ of a target chemical graph $\C$,
 where the bond-multiplicity of each edge in $H^*$ has be determined.
 Finally we introduce a set of rules for constructing 
   a target chemical graph $\C$ from $H^*$ 
   by choosing  a chemical element $\ta\in \Lambda$ 
and assigning a ${\rho}$-fringe-tree $\psi$
 to each interior-vertex $v\in V^\inte$. 
We introduce the following rules for specifying
the size of $\C$, a set of chemical rooted trees  
that are allowed to use as  ${\rho}$-fringe-trees 
and lower and upper bounds on the frequency of
a chemical element, a chemical symbol, % an adjacency-configuration
and an edge-configuration,
where we call the set of prescribed constants   
 a  {\em chemical specification} $\sce$:   
 
\begin{enumerate}[leftmargin=*]
\item[-] 
Lower and upper bounds $n_\LB,  n^*\in \mathbb{Z}_+$
on the number of vertices, where $\nint_\LB \leq n_\LB\leq n^*$.
 
\item[-] 
Subsets  $\mathcal{F}(v) \subseteq \mathcal{F}(D_\pi), v\in \VC$ 
and $\mathcal{F}_E \subseteq \mathcal{F}(D_\pi)$ 
 of chemical rooted trees $\psi$ with $\h(\anpsi)\leq {\rho}$, where 
 we require that 
 every ${\rho}$-fringe-tree $\C[v]$ rooted at a vertex $v\in \VC$
 (resp., at an internal vertex $v$ not in $\VC$)   in  $\C$ 
 belongs to $\mathcal{F}(v)$ (resp.,   $\mathcal{F}_E$).  
Let  $\mathcal{F}^*:=\mathcal{F}_E\cup \bigcup_{v\in \VC}\mathcal{F}(v)$
and 
$\Lambda^\ex$ denote the set of  chemical elements assigned to non-root
vertices over all chemical rooted trees in $\mathcal{F}^*$.  
 
\item[-] 
A subset  $\Lambda^\inte\subseteq \Lambda^\inte(D_\pi)$, where 
 we require that every chemical element $\alpha(v)$ 
 assigned to an interior-vertex  $v$ in $\C$ belongs to $\Lambda^\inte$.
Let $\Lambda:= \Lambda^\inte\cup \Lambda^\ex$ and
 $\na_\ta(\C)$ (resp., $\na_\ta^\inte(\C)$ and $\na_\ta^\ex(\C)$) 
 denote the number of vertices   (resp.,   interior-vertices and  exterior-vertices)
  $v$ such that $\alpha(v)=\ta$   in  $\C$.
 
\item[-] 
A set $\Ldg^\inte\subseteq \Lambda\times [1,4]$  of chemical  symbols
and  a set $\Gamma^\inte \subseteq \Gamma^\inte(D_\pi)$  
of  edge-configurations  $(\mu,\mu' ,m)$ with $\mu \leq \mu'$, where 
 we require that the edge-configuration $\ec(e)$ of an interior-edge $e$ in $\C$ 
 belongs to $\Gamma^\inte$.
We do not distinguish  $(\mu,\mu' ,m)$ and $(\mu' , \mu,m)$.

\item[-] 
Define  $\Gac^\inte$ to be the set of   adjacency-configurations such that  
$\Gac^\inte:=\{(\ta, \tb, m) \mid (\ta d, \tb d',m)\in \Gamma^\inte\}$.   
Let  $\ac_\nu^\inte(\C), \nu\in \Gac^\inte$   
denote  the number of  interior-edges $e$ such that $\ac(e)=\nu$  in $\C$.
  
\item[-] 
 Subsets $\Lambda^*(v)\subseteq \{\ta\in \Lambda^\inte\mid \val(\ta)\geq 2\}$, 
 $v\in \VC$,  
 we require that every chemical element $\alpha(v)$ 
 assigned to   a vertex $v\in  \VC$
 in the seed graph  belongs to $\Lambda^*(v)$.  

\item[-] Lower and upper bound functions 
$\na_\LB,\na_\UB: \Lambda\to  [1,n^*]$  and 
$\na_\LB^\inte,\na_\UB^\inte: \Lambda^\inte\to  [1,n^*]$ 
on the number of   interior-vertices  $v$ such that  $\alpha(v)=\ta$  in $\C$. 

\item[-] Lower and upper bound functions  
$\ns_\LB^\inte,\ns_\UB^\inte: \Ldg^\inte\to  [1,n^*]$ 
  on the number of   interior-vertices $v$ such that $\cs(v)=\mu$  in $\C$.   

\item[-] Lower and upper bound functions  
$\ac_\LB^\inte,\ac_\UB^\inte: \Gac^\inte \to  \mathbb{Z}_+$ 
 on the number of  interior-edges $e$ such that $\ac(e)=\nu$  in $\C$. 

\item[-] Lower and upper bound functions  
$\ec_\LB^\inte,\ec_\UB^\inte: \Gamma^\inte \to  \mathbb{Z}_+$ 
 on the number of  interior-edges $e$ such that $\ec(e)=\gamma$  in $\C$.  
 
 \item[-] Lower and upper bound functions  
$\fc_\LB,\fc_\UB: \mathcal{F}^*\to  [0,n^*]$ 
  on the number of   interior-vertices $v$ 
  such that $\C[v]$ is r-isomorphic to $\psi\in \mathcal{F}^*$  in $\C$.   
  
 \item[-] Lower and upper bound functions  
$\ac^\lf_\LB,\ac^\lf_\UB: \Gac^\lf \to  [0,n^*]$ 
  on the number of  leaf-edges $uv$ in $\acC$
  with adjacency-configuration $\nu$.  
\end{enumerate}
 
We call a chemical graph $\C$ that satisfies a chemical specification $\sce$
a {\em $(\sint,\sce)$-extension of $\GC$},
and denote by $\mathcal{G}(\GC, \sint,\sce)$ the set of
all $(\sint,\sce)$-extensions of $\GC$. 

Table~\ref{table:chemical_spec}  shows an example of 
a chemical-specification  $\sce$ to the seed graph  $\GC$
 in Figure~\ref{fig:specification_example_1}. 
 
%{\tt /* Chemical elements with multi-valence are new */ } %\newone

\begin{table}[h!]\caption{Example~2 of a chemical-specification  $\sce$.  
}
\begin{tabular}{ |  l |  } \hline
 $n_\LB=30$,  $n^* =50$. \\\hline
  branch-parameter:   ${\rho}=2$  \\\hline
\end{tabular}

\begin{tabular}{ |  l |  } \hline
 Each of sets $\mathcal{F}(v), v\in \VC$ and
 $\mathcal{F}_E$ is set to be \\
 the set $\mathcal{F}$  of chemical rooted trees $\psi$ with $\h(\anpsi)\leq {\rho}=2$
in Figure~\ref{fig:specification_example_1}(b). \\\hline
\end{tabular}

\begin{tabular}{ |  c | c |   } \hline
  $\Lambda=\{ \ttH,\ttC,\ttN,\ttO, \ttS_{(2)},\ttS_{(6)}, \ttP=\ttP_{(5)}\}$ & 
  $\Ldg^\inte =\{ \ttC2 , \ttC3,  \ttC4, \ttN2, \ttN3, \ttO2,
    \ttS_{(2)}2,  \ttS_{(6)}3, \ttP4   \}$  
\\\hline
\end{tabular}

\begin{tabular}{ |  c | l |  } \hline
  $\Gac^{\inte}$ &
  $ \nu_1 \!=\!(\ttC   , \ttC  , 1) ,   \nu_2 \!=\!(\ttC   , \ttC  , 2) ,   
   \nu_3 \!=\!(\ttC   , \ttN  , 1) ,  \nu_4 \!=\!(\ttC  , \ttO  , 1), 
    \nu_5 \!=\! (\ttC, \ttS_{(2)}, 1),\nu_6 \!=\!(\ttC  , \ttS_{(6)}, 1), 
    \nu_7 \!=\! (\ttC  , \ttP  , 1) $  \\ \hline
\end{tabular}

\begin{tabular}{ |  c | l |  } \hline
  $\Gamma^{\inte}$ &
  $ \gamma_1 \!=\! (\ttC 2 , \ttC 2, 1) ,
   \gamma_2 \!=\!(\ttC 2 , \ttC 3, 1) ,  
   \gamma_3 \!=\!(\ttC 2 , \ttC 3, 2) ,  
   \gamma_4 \!=\!(\ttC 2 , \ttC 4, 1) , 
   \gamma_5 \!=\!(\ttC 3 , \ttC 3, 1) , 
   \gamma_6 \!=\!(\ttC 3 , \ttC 3, 2) , $ \\
   &
  $   
    \gamma_7 \!=\!(\ttC 3 , \ttC 4, 1), 
   \gamma_8 \!=\!(\ttC 2 , \ttN 2, 1) ,  
   \gamma_9 \!=\!(\ttC 3 , \ttN 2, 1) ,  
   \gamma_{10} \!=\!(\ttC 3 , \ttO 2, 1), 
    \gamma_{11} \!=\!(\ttC 2 , \ttC 2, 2),  
    \gamma_{12} \!=\!(\ttC 2 , \ttO 2, 1) ,$ \\
   &
  $  
    \gamma_{13} \!=\!(\ttC 3 , \ttN3, 1), 
    \gamma_{14} \!=\!(\ttC 4, \ttS_{(2)} 2, 2),  
    \gamma_{15} \!=\!(\ttC 2 , \ttS_{(6)}3, 1), 
   \gamma_{16} \!=\!(\ttC 3 , \ttS_{\tiny (6)}3, 1), 
    \gamma_{17} \!=\!(\ttC 2, \ttP4, 2), $ \\
   &
  $  
    \gamma_{18} \!=\!(\ttC 3, \ttP4, 1)  
     $ \\ \hline
\end{tabular}
    
\begin{tabular}{ |  l|  } \hline
$\Lambda^*(u_1)=\Lambda^*(u_8)=\{{\tt C,  N}\}$, 
$\Lambda^*(u_9)=\{{\tt C, O}\}$, 
   $\Lambda^*(u)=\{\ttC\}$, $u\in \VC\setminus\{u_1,u_8,u_9\}$
   \\\hline
\end{tabular}

\begin{tabular}{ |  c | c c c c  c c c |  } \hline
                         & ${\tt H}$  & ${\tt C}$ &   ${\tt N}$ &     ${\tt O}$ 
                         & $\ttS_{(2)}$ & $\ttS_{(6)}$ & $\ttP$  \\\hline
 $\na_\LB(\ta)$ & 40 &  27 &  1 &   1 & 0 & 0 & 0   \\ \hline 
 $\na_\UB(\ta)$ & 65 & 37 & 4 &  8  &   1 &   1 &   1 \\\hline
\end{tabular} 
\begin{tabular}{ |  c | c c c  c c c   |  } \hline
   & $\ttC$ &   $\ttN$ &     $\ttO$  & $\ttS_{(2)}$ & $\ttS_{(6)}$ & $\ttP$  \\\hline
 $\na_\LB^{\inte}(\ta)$ &   9 &  1 &   0  & 0 & 0 & 0      \\ \hline
 $\na_\UB^{\inte}(\ta) $&  23 & 4 & 5 &   1 &   1 &   1  \\\hline
\end{tabular} 

\begin{tabular}{ |  c | c c c c c c  c c c   |  } \hline
    & $\ttC2$ &  $\ttC3$ &   $\ttC4$ & $\ttN2$ &   $\ttN3$ &   $\ttO2$
   & $\ttS_{(2)}2$ & $\ttS_{(6)}3$ & $\ttP4$  \\\hline
 $\ns_\LB^{\inte}(\mu)$ &  3 &  5 &   0 & 0 &  0 &   0 & 0 &  0 &   0    \\ \hline
 $\ns_\UB^{\inte}(\mu) $&  8 & 15 & 2 & 2 & 3 &  5  &   1 &   1 &   1   \\\hline
\end{tabular} 

\begin{tabular}{ |  c | c c c c c c c |  } \hline
         & $\nu_1 $ &   $\nu_2 $ & $\nu_3 $   & $\nu_4 $
         &   $\nu_5 $ & $\nu_6 $   & $\nu_7 $ \\\hline
 $\ac_\LB^{\inte}(\nu)$  &  0  &  0  & 0  & 0   & 0  & 0 & 0     \\ \hline
 $\ac_\UB^{\inte}(\nu)$ &  30 & 10 & 10 & 10 & 1 & 1 & 1 \\\hline
\end{tabular} 

\begin{tabular}{ |  c | c c c c c c c c c c c c c c c c c c |  } \hline
    & $\gamma_1 $ &   $\gamma_2 $ & $\gamma_3 $   & $\gamma_4 $ 
     & $\gamma_5 $
    & $\gamma_6 $ &   $\gamma_7 $ & $\gamma_8 $   & $\gamma_9 $ 
     & $\gamma_{10} $   & $\gamma_{11} $       & $\gamma_{12} $          
     & $\gamma_{13} $   & $\gamma_{14} $       & $\gamma_{15} $          
     & $\gamma_{16} $   & $\gamma_{17} $       & $\gamma_{18} $          
                            \\\hline
 $\ec_\LB^{\inte}(\gamma)$ &  0 &  0 & 0 &  0  & 0 &  0 &  0 & 0 &  0  & 0 & 0 & 0 
    &  0 & 0 &  0  & 0 & 0 & 0  \\ \hline
 $\ec_\UB^{\inte}(\gamma) $& 4 & 15 & 4 &  4  & 10 &  5 & 4 & 4 &  6 & 4 & 4 & 4
 &  2 & 2 &  2  & 2 & 2 & 2  \\\hline
\end{tabular}

\begin{tabular}{ |  c | c   c   |  } \hline 
& $\psi\in\{\psi_i\mid i=1,6,11\}$ 
& $\psi\in \mathcal{F}^*\setminus \{\psi_i\mid i=1,6,11\}$ \\\hline
 $\fc_\LB(\psi)$  &  1 &    0   \\ \hline 
 $\fc_\UB(\psi)$ &  10 &  3\\\hline
\end{tabular}

\begin{tabular}{ |  c | c   c   |  } \hline 
& $\nu\in\{(\ttC,\ttC,1),(\ttC,\ttC,2)  \}$ 
& $\nu\in \Gac^\lf \setminus \{(\ttC,\ttC,1),(\ttC,\ttC,2)  \}$   \\\hline
 $\ac^\lf_\LB(\nu)$  &  0 &    0   \\ \hline 
 $\ac^\lf_\UB(\nu)$ &  10 &  8 \\\hline
\end{tabular} 

\label{table:chemical_spec}
\end{table}

Figure~\ref{fig:example_chemical_graph} 
 illustrates an example $\Co$ of 
a   $(\sint,\sce)$-extension of $\GC$   obtained 
from the  $\sint$-extension $H^*$  
 in Figure~\ref{fig:specification_example_3} 
under the chemical-specification $\sce$ in Table~\ref{table:chemical_spec}.  
Note that $\mathrm{r}(\Co)= \mathrm{r}(H^*)= \mathrm{r}(\GC)-1=4$
 holds since the edge in $\Ez$ is discarded in $H^*$.

\clearpage

\section{Test Instances for Stages~4 and 5}\label{sec:test_instances} %%%%%%%%%

We prepared the following instances (a)-(d) for conducting experiments
of  Stages~4  and 5 in Phase~2. 
 
 In Stages~4 and 5, we  use three properties 
 $\pi\in \{${\sc OptR}, {\sc SfT}, {\sc Vis}$\}$ 
 and define a set $\Lambda(\pi)$ of chemical elements as follows:  
 $\Lambda(${\sc OptR}$)=\Lambda_5=\{\ttH,\ttC,\ttO, \ttN,\ttS_{(2)},\ttF \}$ and 
   $\Lambda(${\sc SfT}$)=\Lambda(${\sc Vis}$)=\Lambda_4=\{\ttH,\ttC,\ttO, \ttSi_{(4)} \}$. 
 
\begin{itemize} 
  \item[(a)]  $I_{\mathrm{a}} =(\GC,\sint,\sce)$: The instance
  introduced in Section~\ref{sec:specification} to explain the target specification.
For each property $\pi$, we replace
 $\Lambda=\{ \ttH,\ttC,\ttN,\ttO, \ttS_{(2)},\ttS_{(6)},\ttP_{(5)}\}$
in Table~\ref{table:chemical_spec} 
 with $\Lambda(\pi)\cap \{\ttS_{(2)},\ttS_{(6)},\ttP_{(5)}\}$
 and  remove from the $\sce$
 all chemical symbols,  edge-configurations and fringe-configurations
  that cannot be constructed from the replaced element set 
 (i.e., those containing a chemical element in 
 $\{\ttS_{(2)},\ttS_{(6)}, \ttP_{(5)}\}\setminus \Lambda(\pi)$). 
 \end{itemize}

\begin{itemize} 
  \item[(b)] $I_\mathrm{b}^i=(\GC^i,\sint^i, \sce^i)$, $i=1,2,3,4$:
 An instance for inferring chemical graphs with rank at most 2.  
In the four instances $I_\mathrm{b}^i$, $i=1,2,3,4$, 
the following specifications in $(\sint,\sce)$ are common. 
\begin{enumerate}
\item[] 
Set  $\Lambda:=\Lambda(\pi)$
 for a given property $\pi\in \{${\sc   OptR,  SfT, Vis}$\}$, 
 set $\Ldg^\inte$ to be
the set of all possible symbols in $\Lambda\times[1,4]$  
that appear in the data set $D_\pi$  
and set $\Gamma^\inte$
to be the set  of  all  edge-configurations that appear in the data set $D_\pi$. 
Set  $\Lambda^*(v):= \Lambda$,  $v\in \VC$. 
 
\item[] 
The lower bounds  
 $\ell_\LB $, $\bl_\LB $, $\ch_\LB $,  
 $\bd_{2,\LB}$,   $\bd_{3,\LB}$,  
 $\na_\LB$,  $\na^\inte_\LB$,  $\ns^\inte_\LB$,  
$\ac^\inte_\LB$, $\ec^\inte_\LB$ and $\ac^\lf_\LB$  are all set to be 0.

\item[] 
Set  upper bounds   
 $\na_\UB(\ta):=n^*, \na\in\{\ttH,\ttC\}$,   
 $\na_\UB(\ta):=5, \na\in\{\ttO,\ttN\}$,
 $\na_\UB(\ta):=2, \na\in\Lambda\setminus \{\ttH,\ttC,\ttO,\ttN\}$. 
The other upper bounds  
 $\ell_\UB $, $\bl_\UB $, $\ch_\UB $,  
 $\bd_{2,\UB}$,   $\bd_{3,\UB}$,  
% $\na_\UB$, 
 $\na^\inte_\UB$,  $\ns^\inte_\UB$,  
$\ac^\inte_\UB$, $\ec^\inte_\UB$ and $\ac^\lf_\UB$ 
are all set to be an upper bound $n^*$  on $n(G^*)$.

\item[] 
We specify $n_\LB$ as a parameter and
set
$n^*:=n_\LB+10$,
  $\nint_\LB:=\lfloor (1/4) n_\LB \rfloor$ and
   $\nint_\LB:=\lfloor (3/4) n_\LB \rfloor$. 

\item[] 
   For each property $\pi$, let $\mathcal{F}(D_\pi)$ denote
    the set of 2-fringe-trees in the compounds in $D_\pi$,
   and select a subset $\mathcal{F}_\pi^i\subseteq  \mathcal{F}(D_\pi)$ with
   $|\mathcal{F}_\pi^i|=45-5i$, $i\in [1,5]$.
   For each instance $I_\mathrm{b}^i$, 
   set $\mathcal{F}_E :=\mathcal{F}(v):=  \mathcal{F}_\pi^i$, $v\in \VC$ and 
$\fc_\LB(\psi):=0, \fc_\UB(\psi):=10, \psi\in  \mathcal{F}_\pi^i$. 
\end{enumerate}
 
  Instance $I_\mathrm{b}^1$ is given   by the rank-1 seed graph $\GC^1$ 
  in Figure~\ref{fig:specification_example_polymer}(i)
  and   Instances $I_\mathrm{b}^i$, $i=2,3,4$ are
   given by  the rank-2 seed graph $\GC^i$, $i=2,3,4$ in 
   Figure~\ref{fig:specification_example_polymer}(ii)-(iv).

\begin{itemize} 
 \item[(i)]  For instance $I_\mathrm{b}^1$, select as a seed graph 
  the monocyclic graph   $\GC^1=(\VC,\EC=\Et\cup \Ew)$
  in Figure~\ref{fig:specification_example_polymer}(i),
  where $\VC=\{u_1,u_2\}$, $\Et=\{a_1\}$ and  $ \Ew=\{a_2\}$. 
%Set $\nint_\LB:=5, \nint_\UB:=15, n_\LB:=35$ and $n^*:=38$.
We  include a linear constraint 
$\ell(a_1)\leq \ell(a_2)$ 
and $5\leq \ell(a_1)+\ell(a_2) \leq 15$  as part of the side constraint. 
%   $n(G^\dagger)=38, \nint(G^\dagger)=12$.
  
 \item[(ii)]
 For instance $I_\mathrm{b}^2$, select as a seed graph 
  the  graph   $\GC^2=(\VC,\EC=\Et\cup \Ew\cup \Eew)$ 
  in Figure~\ref{fig:specification_example_polymer}(ii),
  where
$\VC=\{u_1,u_2,u_3,u_4\}$, 
$\Et=\{a_1,a_2\}$, 
$\Ew=\{a_3\}$  and 
$\Eew=\{a_4,a_5\}$. 
%Set $\nint_\LB:=25, \nint_\UB:=30, n_\LB:=45$ and $n^*:=50$. 
%
We include a linear constraint $\ell(a_1)\leq \ell(a_2)$ 
and $\ell(a_1)+\ell(a_2)+\ell(a_3)\leq 15$. 
    
 \item[(iii)]
 For instance $I_\mathrm{b}^3$, select as a seed graph 
  the  graph   $\GC^3=(\VC,\EC=\Et\cup \Ew\cup \Eew)$ 
  in Figure~\ref{fig:specification_example_polymer}(iii),   where
$\VC=\{u_1,u_2,u_3,u_4\}$, 
$\Et=\{a_1\}$, 
$\Ew=\{a_2, a_3\}$  and 
$\Eew=\{a_4,a_5\}$. 
%
%Set $\nint_\LB:=25, \nint_\UB:=30, n_\LB:=45$ and $n^*:=50$. 
We include   linear constraints 
$\ell(a_1)\leq \ell(a_2)+\ell(a_3)$, $\ell(a_2)\leq \ell(a_3)$
and $\ell(a_1)+\ell(a_2)+\ell(a_3)\leq 15$.  

 \item[(iv)] 
 For instance $I_\mathrm{b}^4$, select as a seed graph 
  the  graph   $\GC^4=(\VC,\EC=\Et\cup \Ew\cup \Eew)$ 
  in Figure~\ref{fig:specification_example_polymer}(iv),   where
$\VC=\{u_1,u_2,u_3,u_4\}$, 
$\Ew=\{a_1, a_2, a_3\}$  and 
$\Eew=\{a_4,a_5\}$. 
%Set $\nint_\LB:=25, \nint_\UB:=30, n_\LB:=45$ and $n^*:=50$. 
We   include   linear constraints 
$\ell(a_2)\leq \ell(a_1)+1$,
$\ell(a_2)\leq \ell(a_3)+1$,  $\ell(a_1)\leq \ell(a_3)$  
and $\ell(a_1)+\ell(a_2)+\ell(a_3)\leq 15$. 
 \end{itemize}
 \end{itemize}

 We define instances in (c) and (d) 
 in order to find chemical graphs that have an intermediate structure of
 given two chemical cyclic graphs 
 $G_A=(H_A=(V_A,E_A),\alpha_A,\beta_A)$ 
and $G_B=(H_B=(V_B,E_B),\alpha_B,\beta_B)$.
Let
 $\Lambda_A^\inte$ and  $\Lambda_{\mathrm{dg},A}^\inte$ 
 denote the sets  of chemical elements
 and chemical symbols  of
 the interior-vertices in $G_A$, 
 $\Gamma_A^\inte$   denote the sets of edge-configurations of
  the interior-edges in $G_A$,   
  and 
  $\mathcal{F}_A$ denote the set of 2-fringe-trees in $G_A$.  
Analogously define sets
 $\Lambda_B^\inte$,    $\Lambda_{\mathrm{dg},B}^\inte$,   
 $\Gamma_B^\inte$ and   $\mathcal{F}_B$ 
 in $G_B$.

\begin{itemize}  
\item[(c)]  $I_{\mathrm{c}}=(\GC,\sint,\sce)$: 
An instance aimed to infer a chemical graph $G^\dagger$ such that
the core of $G^\dagger$ is equal to the core of $G_A$ and 
the frequency of each edge-configuration in the non-core of $G^\dagger$
is equal to that of  $G_B$. 
We use chemical compounds CID~24822711 and CID~59170444 in 
 Figure~\ref{fig:instance_I_c_I_d}(a) and (b)
 for $G_A$ and $G_B$, respectively.  \\
% Assume that $\Lambda_A^\inte$ contains  any chemical element  of
% a core-vertex  adjacent to  non-core-vertices in $G_B$;  \\ 
Set   a seed graph $\GC=(\VC,\EC=\Eew)$ to be the core of $G_A$. \\
Set  $\Lambda:=\{{\tt H,C,N,O}\}$, 
and  set $\Ldg^\inte$ to be
the set of all possible chemical symbols in $\Lambda\times[1,4]$.\\
Set 
$\Gamma^\inte:=\Gamma_A^\inte\cup \Gamma_B^\inte$ and 
  $\Lambda^*(v):=\{\alpha_A(v)\}$, $v\in \VC$.  \\
Set 
$\nint_\LB:=\min\{\nint(G_A), \nint(G_B)\}$, 
$\nint_\UB:=\max\{\nint(G_A), \nint(G_B)\}$, \\
$n_\LB:=\min\{n(G_A), n(G_B)\}-10=40$ 
and   $n^*:=\max\{n(G_A), n(G_B)\}+5$. \\
Set  lower bounds  
 $\ell_\LB $, $\bl_\LB $, $\ch_\LB $,  
 $\bd_{2,\LB}$,   $\bd_{3,\LB}$,  
 $\na_\LB$,  $\na^\inte_\LB$,  $\ns^\inte_\LB$, 
$\ac^\inte_\LB$ and  $\ac^\lf_\LB$  to be 0.\\
Set  upper bounds   
 $\na_\UB(\ta):=n^*, \na\in\{\ttH,\ttC\}$,   
 $\na_\UB(\ta):=5, \na\in\{\ttO,\ttN\}$,
 $\na_\UB(\ta):=2, \na\in\Lambda\setminus \{\ttH,\ttC,\ttO,\ttN\}$ 
and set the other upper bounds
 $\ell_\UB $, $\bl_\UB $, $\ch_\UB $,  
 $\bd_{2,\UB}$,   $\bd_{3,\UB}$,  
% $\na_\UB$,  
$\na^\inte_\UB$,  $\ns^\inte_\UB$, 
$\ac^\inte_\UB$   and  $\ac^\lf_\UB$ to be  $n^*$. \\
Set $\ec_\LB^\inte(\gamma)$ 
to be the number of core-edges  in $G_A$ with $\gamma\in \Gamma^\inte$ and  
 $\ec_\UB^\inte(\gamma)$  
to be the number interior-edges in $G_A$ and  $G_B$ 
with edge-configuration $\gamma$. \\
Let $\mathcal{F}_B^{(p)}, p\in [1,2]$ denote the set of chemical rooted 
trees r-isomorphic $p$-fringe-trees in $G_B$; \\
Set $\mathcal{F}_E :=\mathcal{F}(v):= 
 \mathcal{F}_B^{(1)}\cup \mathcal{F}_B^{(2)}$, $v\in \VC$ and
$\fc_\LB(\psi):=0, \fc_\UB(\psi):=10, \psi\in \mathcal{F}_B^{(1)}\cup \mathcal{F}_B^{(2)}$. 
 
  \item[(d)] $I_{\mathrm{d}}=(\GC^1,\sint, \sce)$:     
An instance aimed to infer a chemical monocyclic graph $G^\dagger$ such that
the frequency vector of  edge-configurations in  $G^\dagger$
is a vector obtained by merging those of $G_A$ and $G_B$.
We use chemical monocyclic compounds CID~10076784 and CID~44340250
in   Figure~\ref{fig:instance_I_c_I_d}(c) and (d) 
 for $G_A$ and $G_B$, respectively.  
%Assume that $G_A$ and $G_B$ are monocyclic.
Set a seed graph to be   the monocyclic seed graph  
 $\GC^1=(\VC,\EC=\Et\cup \Ew)$ with 
  $\VC=\{u_1,u_2\}$, $\Et=\{a_1\}$ and  $ \Ew=\{a_2\}$ 
  in Figure~\ref{fig:specification_example_polymer}(i). \\
Set  $\Lambda:=\{{\tt H,C,N,O}\}$,  
 $\Ldg^\inte:=\Lambda_{\mathrm{dg},A}^\inte 
                 \cup \Lambda_{\mathrm{dg},B}^\inte$ and 
$\Gamma^\inte:=\Gamma_A^\inte\cup \Gamma_B^\inte$. \\
Set 
$\nint_\LB:=\min\{\nint(G_A), \nint(G_B)\}$, 
$\nint_\UB:=\max\{\nint(G_A), \nint(G_B)\}$, \\
  $n_\LB:=\min\{n(G_A),n(G_B)\}=40$ and  
  $n^*:=\max\{n(G_A),n(G_B)\}$. \\
Set  lower bounds  
 $\ell_\LB $, $\bl_\LB $, $\ch_\LB $,  
 $\bd_{2,\LB}$,   $\bd_{3,\LB}$,  
 $\na_\LB$,  $\na^\inte_\LB$,  $\ns^\inte_\LB$, 
$\ac^\inte_\LB$  and  $\ac^\lf_\LB$ to be 0.\\
Set  upper bounds   
 $\na_\UB(\ta):=n^*, \na\in\{\ttH,\ttC\}$,   
 $\na_\UB(\ta):=5, \na\in\{\ttO,\ttN\}$,
 $\na_\UB(\ta):=2, \na\in\Lambda\setminus \{\ttH,\ttC,\ttO,\ttN\}$ 
and set the other  upper bounds  
 $\ell_\UB $, $\bl_\UB $, $\ch_\UB $,  
 $\bd_{2,\UB}$,   $\bd_{3,\UB}$,  
%$\na_\UB$,
  $\na^\inte_\UB$,  $\ns^\inte_\UB$,
$\ac^\inte_\UB$ and  $\ac^\lf_\UB$  to be   $n^*$. \\
For each edge-configuration
 $\gamma \in \Gamma^\inte$,  
let  $\x^*_A(\gamma^\inte)$  (resp., $\x^*_B(\gamma^\inte)$)   denote
 the number of interior-edges with $\gamma$ in $G_A$ (resp., $G_B$), 
 $\gamma \in \Gamma^\inte$ and   
set \\
$\x^*_{\min}(\gamma):=\min\{\x^*_A(\gamma), \x^*_B(\gamma)\}$, 
 $\x^*_{\max}(\gamma):=\max\{\x^*_A(\gamma), \x^*_B(\gamma)\}$, \\
$\ec_\LB^\inte(\gamma):=
\lfloor (3/4)\x^*_{\min}(\gamma)+(1/4)\x^*_{\max}(\gamma) \rfloor$
and  \\
$\ec_\UB^\inte(\gamma):=
\lceil (1/4)\x^*_{\min}(\gamma)+(3/4)\x^*_{\max}(\gamma) \rceil$. \\
Set $\mathcal{F}_E :=\mathcal{F}(v):=  \mathcal{F}_A\cup \mathcal{F}_B$, 
$v\in \VC$ and 
$\fc_\LB(\psi):=0, \fc_\UB(\psi):=10, \psi\in \mathcal{F}_A\cup \mathcal{F}_B$. \\
We  include a linear constraint 
$\ell(a_1)\leq \ell(a_2)$ 
and $5\leq \ell(a_1)+\ell(a_2) \leq 15$  as part of the side constraint. 
 \end{itemize}

\end{document}